\documentclass[review,1p,times,numbers]{elsarticle}
\usepackage{tabulary}
\usepackage{amsfonts,amsmath,amssymb}
\usepackage[T1]{fontenc}
\usepackage{natbib}
\usepackage[utf8]{inputenc}
\usepackage{stmaryrd}
\usepackage{multirow,morefloats,floatflt,cancel,tfrupee}
\usepackage{colortbl}
\usepackage{pifont}
\usepackage[nointegrals]{wasysym}
\usepackage{float}
\usepackage[labelfont=bf,font={small,color=Cadet}]{caption} 
\usepackage{longtable}
\usepackage[algo2e, algoruled, algosection, linesnumbered, noend]{algorithm2e}
\usepackage[hyphens]{url}
\usepackage[bookmarks=true, pdfstartview=Fit, linktoc=page, pdfpagemode=UseNone]{hyperref}
\usepackage[usenames,dvipsnames,hyperref]{xcolor}
\definecolor{Cadet}{rgb}{0.33, 0.41, 0.47}
\definecolor{beaublue}{rgb}{0.74, 0.83, 0.9}
\definecolor{Asparagus}{rgb}{0.53, 0.66, 0.42}
\definecolor{table_title}{HTML}{79A8A4}
\definecolor{table_row_highlight}{HTML}{E9ECE6}
\newenvironment{SgAlgorithm}[1][t]
{%
	\begin{algorithm2e}[#1]
    \linespread{1.2}
    \selectfont
}
{\end{algorithm2e}}

\definecolor{mygreen}{HTML}{E9ECE6}
\usepackage{amsmath}
\DeclareMathOperator*{\argmin}{arg\,min} 

\begin{document}
\begin{frontmatter}
    \title{Taxlifier: Leveraging Disease Taxonomy for Enhanced Multi-Label Classification in Chest Radiography}
    \author[]{Mohammad S\@. Majdi}
    \author[]{Jeffrey J\@. Rodriguez\corref{co1}}
    \affiliation[]{organization={Dept\@. of Electrical and Computer Engineering, The University of Arizona}, city={Tucson}, postcode={85721}, state={AZ}, country={USA}}
    \begin{abstract}
Accurate and efficient classification of thoracic diseases in chest X-ray (CXR) images is crucial for timely diagnosis and treatment. However, the presence of multiple pathologies with overlapping visual characteristics poses significant challenges for automated classification systems. In this study, we propose two novel hierarchical multi-label classification techniques, namely the loss-based and logit-based methods, to address these challenges by leveraging the hierarchical relationships among different thoracic pathologies. The loss-based technique integrates hierarchical information directly into the optimization process, while the logit-based method adjusts the predicted probabilities of each class based on its parent class in the disease taxonomy. We evaluate the performance of both techniques using three large-scale CXR datasets: CheXpert (224,316 CXRs), PADCHEST (160,000 CXRs), and NIH (112,120 CXRs). The experimental results demonstrate significant improvements in accuracy, AUC, and F1 scores compared to the baseline method across various pathologies. The logit-based and loss-based methods improve accuracy by 12\% and 11\%, AUC by 13\% and 10\%, and F1 scores by 24\% and 12\%, respectively compared to the baseline. These results represent a substantial improvement over the baseline method. Furthermore, we conduct a comprehensive statistical analysis to validate the robustness and reliability of the proposed techniques. The integration of domain-specific hierarchical knowledge not only enhances the classification performance but also provides a more interpretable output for clinical decision support. Our findings highlight the potential of hierarchical multi-label classification in advancing computer-aided diagnosis systems for chest radiography.
    \end{abstract}
\end{frontmatter}
\section{Introduction}\label{sec:taxonomy.introduction}
Chest X-ray (CXR) is a prevalent radiological examination for diagnosing lung and heart disorders, constituting a significant share of ordered imaging studies. Fast and accurate detection of different thoracic diseases, such as pneumothorax, is crucial for optimal patient care~\cite{bellaviti_Increased_2016}. However, interpreting CXRs can be challenging due to similarities between different thoracic diseases, which may result in misinterpretation even by experienced radiologists~\cite{delrue_Difficulties_2011}. Consequently, devising an accurate system to identify and localize common thoracic diseases can aid radiologists in minimizing diagnostic errors~\cite{crisp_Global_2014,silverstein_Most_2016}.

Progress in natural language processing (NLP) has enabled the collection of extensive annotated datasets such as ChestX-ray8~\cite{wang_ChestXRay8_2017}, PADCHEST~\cite{bustos_Padchest_2020}, and CheXpert~\cite{irvin_CheXpert_2019}, allowing researchers to develop more efficient and robust supervised learning algorithms. Convolutional neural networks (CNNs) exhibit potential for learning intricate relationships between image objects. However, their training necessitates vast amounts of labeled data, which can be both expensive and time-consuming to acquire. Despite these challenges, deep learning techniques have become increasingly popular in medical imaging, especially in radiology, due to their ability to perform complex tasks with minimal human intervention~\cite{jaderberg_Spatial_2015}.

The timely diagnosis and effective treatment of diseases depend on the fast and accurate detection of anomalies in medical images. Deep learning techniques have made substantial progress in the medical imaging domain, exhibiting impressive success across various applications~\cite{litjens_Survey_2017a,eshghali_Machine_2023}.  Although recent advances in deep learning have facilitated the creation of CAD systems capable of classifying and localizing prevalent thoracic diseases using CXR images, most of these techniques have concentrated on specific diseases~\cite{jaiswal_Identifying_2019,lakhani_Deep_2017,pasa_Efficient_2019,ausawalaithong_Automatic_2018}, leaving ample opportunities to investigate a unified deep learning framework that can efficiently detect a broad spectrum of common thoracic diseases.
Furthermore, conventional multi-label classification methods are primarily designed for single-label predictions and struggle with capturing intricate label relationships~\cite{tsoumakas_MultiLabel_2007}. Many existing approaches, such as the One-vs-All (OVA) method, treat each label independently, ignoring potential dependencies between labels. This can lead to suboptimal performance, especially when dealing with hierarchical label structures common in medical taxonomies. Moreover, labels at the lower levels of the hierarchy, particularly leaf nodes, often have very few positive examples, making flat classification models susceptible to negative class bias.
While some existing works have tackled the idea of incorporating taxonomical information in classification tasks, they are mainly focused on either text applications~\cite{pourvali_taxoknow_2023} or single-class multitask problems~\cite{bjerge_hierarchical_2023,ho_hierarchical_2023,ma_multitask_2021}. However, there is a lack of research on leveraging taxonomical information for multi-class problems in imaging applications.
To address these challenges, we propose a hierarchical multi-label classification framework that incorporates the relationships between different classes to provide more accurate and interpretable predictions. Our framework makes two key methodological contributions:
\begin{enumerate}
    \item A loss-based approach for scenarios where ground truth is available, which integrates hierarchical label relationships directly into the loss function (e.g., a classification or segmentation network such as DenseNet121~\cite{huang_Densely_2017} or U-Net~\cite{ronneberger_UNet_2015}). This is achieved by introducing a novel hierarchical regularization term that penalizes misalignment with the label taxonomy during training.
    \item A logit-based approach for scenarios where ground truth is unavailable, which adjusts the logit outputs of a pre-trained neural network based on class relationships in a computationally efficient post-processing step. This enables leveraging taxonomical knowledge without requiring extensive modifications to the model architecture or expensive retraining.
\end{enumerate}
Both techniques allow incorporating label hierarchy information into existing multi-label classification frameworks with minimal modifications, making them easily adaptable to a wide range of models and applications.
In summary, the main contributions of this work are:
\begin{itemize}
    \item A loss-based technique that directly incorporates label hierarchy into the model training process via a novel regularization term
    \item A logit-based technique that adjusts neural network outputs to account for label relationships in a computationally efficient post-processing step
    \item Extensive evaluation on three public chest X-ray datasets demonstrating consistent improvements in classification performance metrics
    \item Enabling multi-label classification models to leverage taxonomical knowledge for improved accuracy and reduced negative class bias
\end{itemize}
The remainder of this paper is structured as follows: Section~\ref{sec:taxonomy.relatedwork} discusses related work on multi-label classification and hierarchical loss functions; Section~\ref{sec:taxonomy.methods} describes the proposed techniques for integrating label hierarchy into multi-label classification; Section~\ref{sec:taxonomy.results} presents experimental results using chest radiograph datasets; and Section~\ref{sec:taxonomy.discussion} concludes the paper and outlines future research directions.

\section{Related Work}\label{sec:taxonomy.relatedwork}
The introduction of the ChestX-ray8 dataset and its associated model by Wang et al.\cite{wang_ChestXRay8_2017} marked a significant milestone in large-scale chest radiograph (CXR) classification, leading to numerous advancements in both modeling techniques and dataset collection. Subsequent improvements have encompassed the integration of ensemble methods\cite{islam_Abnormality_2017}, attention mechanisms~\cite{guan_Diagnose_2018,liu_SDFN_2019}, and localization techniques~\cite{cai_Iterative_2018,guendel_MultiTask_2019,li_Thoracic_2018,yan_Weakly_2018}. Early approaches frequently employed "binary relevance" (BR) learning, which reduces the multi-label classification problem to a series of binary classification tasks by training a separate binary classifier for each class~\cite{zhang_Review_2014}. However, BR-based techniques fail to capture label dependence, which can manifest as either conditional (instance-specific) or marginal (dataset-specific) dependence~\cite{dembczynski_Label_2012}.

Multi-label classification poses unique challenges compared to multi-class methods, as it necessitates the simultaneous classification of instances into multiple categories. For instance, a single CXR image may exhibit both edema and cardiomegaly~\cite{harvey_Standardised_2019,tsoumakas_MultiLabel_2007}. Prior to the widespread adoption of deep learning, considerable research efforts were directed towards integrating taxonomies through hierarchical classification by extracting binary hierarchical multi-label classification (HMLC) labels from pseudo-probability predictions~\cite{bi_BayesOptimal_2015}. Early methods relied on hierarchical and multi-label extensions of traditional algorithms, such as nearest-neighbor, multi-layer perceptron~\cite{pourghassem_ContentBased_2008}, and decision trees~\cite{dimitrovski_Hierarchical_2011}. With the emergence of deep learning, the adaptation of convolutional neural networks (CNNs) for hierarchical classification has garnered increasing attention~\cite{guo_CNNRNN_2018,kowsari_HDLTex_2017,redmon_YOLO9000_2017,roy_TreeCNN_2020}.

In the context of medical imaging, the diagnosis or observation of a particular condition on a CXR is often influenced by the presence or absence of the parent class~\cite{vaneeden_Relationship_2012}. For example, when diagnosing pneumonia, a radiologist may first seek evidence of lung consolidation (parent label) in the CXR. Considering the relationships between labels can lead to more accurate diagnoses. However, many existing CXR classification methods treat each label independently, adopting "flat classification" approaches~\cite{alaydie_Exploiting_2012}. Moreover, labels at lower levels of the hierarchy, particularly leaf nodes, often have very few positive examples, rendering flat learning models susceptible to negative class bias. Addressing these challenges necessitates the development of models that consider the hierarchical nature of CXR data.

Hierarchical multi-label classification methods have been successfully applied in various domains, such as text processing~\cite{aly_Hierarchical_2019} and genomic analysis~\cite{bi_BayesOptimal_2015}. Rana~\cite{rana_weakly_2023} introduced a weakly supervised Hierarchical Multi-task Classification Framework (HMCF) for identifying topics from customer questions in e-commerce websites at various granularities. A common technique~\cite{chen_Deep_2019} for leveraging hierarchies involves training a classifier on conditional data while excluding samples with negative parent-level labels, followed by reintroducing these samples to fine-tune the network across the entire dataset. Although effective, these approaches are computationally demanding, making their application to large-scale real-world datasets challenging. An alternative strategy is to employ a cascading architecture with separate classifiers trained at each level of the hierarchy, but this also requires substantial computational resources. Existing deep learning-based approaches often rely on complex combinations of CNNs and recurrent neural networks (RNNs)~\cite{guo_CNNRNN_2018,kowsari_HDLTex_2017}.

While these studies have made valuable contributions to the field of hierarchical classification, they primarily focus on single-class multitask problems~\cite{bjerge_hierarchical_2023} or text-based applications~\cite{aly_Hierarchical_2019;rana_weakly_2023}. In contrast, our research addresses the specific challenges of multi-class problems in medical imaging. To the best of our knowledge, there is limited research on leveraging taxonomical information for multi-class problems in the context of medical image analysis, underscoring the novelty and significance of our proposed approach.

In this study, we introduce two innovative hierarchical multi-label classification techniques that aim to improve the accuracy and interpretability of results in applications with hierarchical class structures. By incorporating label hierarchy directly into the model's optimization process or output adjustment, our methods offer a novel perspective on addressing the challenges associated with multi-label classification in medical imaging. The proposed techniques are designed to enhance classification performance, increase robustness to labeling inaccuracies, and provide a higher level of alignment with hierarchical class structures. Through extensive evaluations on multiple large-scale chest X-ray datasets, we demonstrate the effectiveness and generalizability of our approach, highlighting its potential for real-world application in the domain of multi-label classification for medical imaging.
\section{Methods}\label{sec:taxonomy.methods}
In this study, we introduce two novel methods to improve the accuracy and interpretability of multi-label classification, with potential applications in areas such as chest radiography.
Two key novel methodological contributions include:
1.	A loss-based approach that integrates hierarchical label relationships directly into the loss function during training by introducing a novel hierarchical regularization term. This term penalizes misalignment with the label taxonomy in a new way.
2.	A logit-based approach that adjusts logit outputs based on class relationships in a computationally efficient post-processing step, enabling the incorporation of taxonomical knowledge without requiring extensive model modifications or retraining, unlike most previous approaches.
One notable advantage of our proposed techniques lies in enhancing interpretability. By categorizing classes into a hierarchical structure and capitalizing on their relationships, the model not only improves classification performance but also provides insights into the relationships among predicted classes.  This additional layer of interpretability can help radiologists in understanding the reasoning behind the model's predictions, fostering trust in the model's output and facilitate its integration into clinical workflows. Furthermore, the hierarchical nature of the taxonomy allows radiologists to explore predictions at various levels of granularity, depending on the level of detail required for a specific case. Furthermore, the proposed technique is adaptable to the available computational resources. When ample computational resources are available, the ``loss-based'' strategy can be utilized. Alternatively, in scenarios with limited computational resources, to avoid the need for optimization of the network from scratch, the ``logit-based'' strategy can be utilized.

\subsection{Problem Formulation}\label{subsec:taxonomy.problem_formulation}

\subsubsection{Mathematical Formulation of Sigmoid Function}
In the context of neural networks, a logit refers to the raw, unscaled output of a neuron. This output is obtained at the last layer of a neural network model prior to the application of the sigmoid layer~\cite{furnieles_Sigmoid_2022}. Logit values can range from negative to positive infinity. The term logit-based originally comes from logistic regression, and it is the inverse of the logistic sigmoid function. In machine learning, it's often desirable for our model to produce real numbers ranging from 0 to 1. Applying the sigmoid function to the logit ensures this, as the sigmoid function maps any real number to the interval \([0,1]\).
The equation representing the sigmoid function is:
\begin{equation}
    p = \text{{sigmoid}}(q) = \frac{1}{1 + e^{-q}}
\end{equation}
When we apply this sigmoid function to the logit values produced by the neural network, the result is a predicted probability ranging from 0 to 1. This property is particularly useful in binary classification tasks, where the aim is to model the probability of a given input pertaining to a certain class.
In a binary classification scenarios, if we apply the sigmoid function to the logit value and obtain output \( p \), we interpret this as the model's estimated probability that the input belongs to the class.
Finally, the equation for the logit (also known as the log-odds) can be given as
\begin{equation}
    q = \text{{logit}}(p) = \log \left( \frac{p}{1 - p} \right)
\end{equation}
where \( p \) is the probability of a positive event. This function maps a probability \( p \) from the interval \((0,1)\) to any real number.

\subsubsection{Glossary of Symbols}\label{subsubsec:notations}
Let us define the following parameters:
\begin{itemize}
    \item  $\mathcal{C} = {\{c_k\}}_{k=1}^{K}  $: the set of classes (categories) in the multi-label dataset, where $c_k $ is the name of the $k $-th class.
    \item  $\mathcal{E} $: set of edges representing parent-child relationships between classes.
    \item  $\mathcal{G}=\left\{\mathcal{C},\mathcal{E}\right\} $:  Graph representing the taxonomy of thoracic diseases.
    \item  $c_j=\Lambda (c_k) \in \mathcal{C}$: parent class of class $c_k $ in graph $\mathcal{G} $.
    \item  $\mathcal{J}(c_j) \subset \mathcal{C}$: set of child classes of class $c_j$ in graph $\mathcal{G} $
    \item  $y_k^{(i)} \in \{0,1\} $: true label for the $k $-th class of instance $i $.
    \item  $q_k^{(i)} \in \left( -\infty,\infty \right) $: logits obtained in the last layer of the neural network model before the sigmoid layer.
    \item  $p_k^{(i)} = \text{sigmoid}\left(q_k^{(i)}\right) = \frac{1}{1+\exp{\left(-q_k^{(i)}\right)}} $: predicted probability for the $k $-th class ($c_k) $ of instance $i $ with a value between 0 and 1. $p_k^{(i)} $ represents the likelihood that class $k $ is present in instance $i $ and is obtained by passing logits $q_k^{(i)} $ through a sigmoid function.
    \item  $\theta_k $: Binarization threshold for class $k $.  To obtain this, we can utilize any existing thresholding technique (for example, in one technique, we analyze the ROC curve and find the corresponding threshold where the difference between the true positive rate (sensitivity) and false positive rate (1-specificity) is maximum; alternatively, we could simply use $0.5 $).
    \item  $t_k^{(i)}=\left\{\begin{array}{ll}1&\text{if}\;p_k^{(i)} \geq \theta_k\\0&\text{otherwise.}\end{array}\right. $: predicted label obtained by binarizing the $p_k^{(i)} $
    \item  ${\widehat p}_k^{(i)} \in (0,1) $: updated predicted probability for the $k $-th class of instance $i $ with a value between 0 and 1.
    \item  $\widehat{t}_k^{(i)}=\left\{\begin{array}{ll}1&\text{if}\;\widehat{p}_k^{(i)}\geq\theta_k\\0&\text{otherwise.}\end{array}\right. $: updated predicted label for the $k $-th class of instance $i $.
    \item $K $: number of categories (aka classes) in a multi-class, multi-label problem. For example, suppose that we have a dataset that is labeled for the presence of cats, dogs, and rabbits in any given image. If a given image $X^{(i)} $ has cats and dogs but not rabbits, then $Y^{(i)} = \{1,1,0\} $.
    \item $N $: Number of instances.
    \item $X^{(i)} $: Data for instance $i$.
    \item $Y^{(i)}=\left\{y_1^{(i)},y_2^{(i)},\;\dots,y_{K}^{(i)}\right\} $: True label set for instance $i $. For example, consider a dataset that is labeled for the presence of cats, dogs, and rabbits in any given instance. If a given instance $X^{(i)} $ has cats and dogs but not rabbits, then $Y^{(i)}=\{1,1,0\} $.
    \item $P^{(i)} = {\left\{ p_k^{(i)} \right\}}_{k=1}^{K} $: Predicted probability set obtained in the output of the classifier $F(\cdot) $ representing the probability that each class $k $ is present in the sample.
    \item $T^{(i)} = {\left\{t_k^{(i)}\right\}}_{k=1}^{K} $: predicted label set for instance $i $.
    \item $\mathbb{X} = {\left\{X^{(i)}\right\}}_{i=1}^{N} $: Set of all instances.
    \item $\mathbb{Y} = {\left\{Y^{(i)}\right\}}_{i=1}^{N} $: Set of all true labels.
    \item $\mathbb{D}=\left\{\mathbb{X},\mathbb{Y}\right\} $: Dataset containing all instances and all true labels.
    \item $l_k^{(i)} = \mathcal{L} \left(y_k^{(i)},p_k^{(i)}\right) $:  $\mathcal{L}( \cdot) $ is an arbitrary loss function (e.g., binary cross entropy) that takes the true label $y_k^{(i)}$ and predicted probability $p_k^{(i)}$ for class $k$ and instance $i$ and outputs the loss value $l_k^{(i)} $. We refer to this as the ``base loss function'' throughout this paper.
    \item $\text{Loss}(\theta) $: Measured loss for all classes and instances. This value is obtained using a modified version of the base loss function $\mathcal{L}(\cdot) $ (e.g., with added regularization, etc.).
    \item $\omega_k^{(i)} $: Estimated weight for $k$-th class $c_k $ of instance $i $ with respect to its parent class $c_j=\Lambda (c_k) $.
    \item ${\widehat l}_k^{(i)} = \omega_k^{(i)} \; l_k^{(i)} $: updated loss for class $k $ and instance $i $.
\end{itemize}
Let us define the multi-label classification problem as follows. Let $\mathbb{X} = {\left\{X^{(i)}\right\}}_{i=1}^{N} $ be a set of $N $ chest radiograph images and $\mathbb{Y} = {\left\{Y^{(i)}\right\}}_{i=1}^{N} $ be their corresponding ground truth labels. The ground-truth labels for the dataset were provided by experienced radiologists who annotated each image with the corresponding abnormalities.

Given the set of disease classes $\mathcal{C} = \{c_1,c_2,\dots,c_K\} $, let us define a  graph $\mathcal{G}=\left\{\mathcal{C},\mathcal{E}\right\} $ representing the taxonomy of thoracic diseases, where $\mathcal{E}$ is the set of edges representing parent-child relationships between these classes. For each node $c_k \in \mathcal{C} $, let $\Lambda (c_k)$ be the parent node of class $c_k $ and let $\mathcal{J}(c_k) \subset \mathcal{C} $ be the set of child classes of class $c_k $ in graph $\mathcal{G}$.

In the context of multi-label classification problems, each sample may have multiple labels assigned to it simultaneously. To this end, we use a deep neural network, with multiple hidden layers and the sigmoid activation function in the final layer. Let's denote the input to this neural network by $x^{(i)}$, which represents data for instance $i$ (data type can be a 1D feature vector, 2D image, or 3D volume). This network is trained to predict the probabilities for each class being present in a given sample. Hence, the output of the final layer of the neural network for instance $i$ is passed through a sigmoid function to generate a set of values, each ranging from 0 to 1, corresponding to the label set $\mathcal{C} $.
The outcome of this operation is a set of $K $ predicted probabilities $P^{(i)}={\left\{p_k^{(i)}\right\}}_{k=1}^{K} $. Each of these predicted probabilities, derived from the sigmoid activation function, can be interpreted as the likelihood that the input sample belongs to each class.

Furthermore, let $\omega_k^{(i)} $ be a scalar weight assigned to the class $c_k $ of instance $i $ with respect to its parent class $\Lambda (c_k)$.
Each of these predicted probabilities, derived from the sigmoid activation function, can be interpreted as the likelihood that the input sample belongs to each class. A loss function is utilized to quantify the similarity between predicted probabilities and true labels. This function guides the learning process of the neural network by providing a measure of the prediction error, which is minimized during the training phase.
Let us denote the loss value as $l_k = \mathcal{L} \left(y_k^{(i)}, p_k^{(i)}\right),\hspace{0.33em}k \in \{1,2,\dots,K\} $ where $\mathcal{L}(\cdot) $ is an appropriate single-class loss function for the task (e.g., binary cross-entropy, Dice, etc.) that is used to calculate the difference between the predicted probability $p_k^{(i)} $ and the true class label $y_k^{(i)} $ for instance $i$ and class $k $.

\subsection{Label Taxonomy Structure}\label{subsec:label-taxonomy-and-hierarchy}
To exploit the hierarchical relationships between thoracic abnormalities, the first step is to define a disease taxonomy that demonstrates different abnormalities' interrelationships. In this taxonomy, diseases are structured hierarchically in a graph, with higher levels representing broader disease categories and lower levels representing more nuanced distinctions between related diseases. The taxonomy is structured such that if a disease is present then its parent disease is also present. Furthermore, in the presence of multiple parent classes for a given child class, the taxonomy structure only utilizes the more dominant parent (e.g., if class $c_1$ has two parent classes $c_3$ and $c_5$, while the $c_5$ is also the parent of $c_3$ class (it's both the parent and grandparent of $c_1$ class), in this scenario, we assume $c_5$ as the parent class of both $c_1$ and $c_3$).

For example, pleural effusion and pneumothorax can be classified as subcategories of pleural abnormalities, whereas atelectasis and consolidation can be classified under pulmonary opacity~\cite{irvin_CheXpert_2019}. This hierarchical structure enables the model to take advantage of the relationships between diseases to improve its classification performance.
In medical imaging, classes are frequently organized as graphs to represent the hierarchical relationships between different classes. For example, a graph can be used to represent the human body's organs, with each node representing a different organ and the edges representing the relationships between organs (e.g., the liver is part of the abdominal cavity). Using a graph structure for labels in medical imaging has a number of advantages, including improved accuracy and interpretability of classification algorithms, which are essential for making sense of the vast amounts of data generated by medical imaging technologies. In medical imaging, hierarchies of labels are typically constructed by subject matter experts with a comprehensive understanding of human anatomy and physiology, such as radiologists.

To create the label taxonomy shown in Figure~\ref{fig:taxonomy.fig.1.taxonomy_structure}, we combined the taxonomies provided by Irvin~\cite{irvin_CheXpert_2019} for the CheXpert dataset, Chen~\cite{chen_Deep_2020} for the PADCHEST~\cite{bustos_Padchest_2020} and the CXR portion of the prostate, lung, colorectal and ovarian (PLCO) dataset~\cite{gohagan_Prostate_2000}.
In order to maintain uniformity, we adopted the renaming scheme introduced by Cohen~\cite{cohen_TorchXRayVision_2022} for the pathology names. Subsequently, the key pathologies were identified and extracted to build the hierarchical taxonomy structure illustrated in Figure~\ref{fig:taxonomy.fig.1.taxonomy_structure}.

\subsection{Approach 1: Conditional Predicted Probability}\label{subsec:taxonomy.method.approach1}
The logit-based approach presents a novel way to incorporate taxonomical knowledge into existing multi-label classification models without requiring extensive modifications or retraining. This is a key advantage over prior hierarchical classification methods, which often rely on complex custom architectures. When computational resources are limited, this technique can be applied to test samples without the need to fine-tune the pre-trained, multi-label classification model. This adaptability ensures that the benefits of considering hierarchical relationships between labels can be realized in a wide range of practical scenarios, without imposing excessive computational requirements.

The proposed logit-based approach offers several potential benefits over existing methods:
\begin{itemize}
\item  \textbf{Simplicity and ease of implementation:} The direct modification of predicted probabilities eliminates the need for substantial changes to the loss function, making the logit-based approach easier to implement compared to methods that require complex modifications to the model architecture or training process.
\item  \textbf{Improved performance in specific scenarios:} Depending on the problem and dataset, the proposed logit-based approach may provide superior performance in certain circumstances, especially when incorporating class relationships into the loss function is challenging or computationally expensive.
\item  \textbf{Easier calibration of model outputs:} The direct modification of predicted probabilities facilitates the calibration of model outputs to more closely match the true label distribution, which is a novel advantage over existing hierarchical classification methods that may require additional calibration steps.
\end{itemize}
The proposed technique provides an easy way to improve the performance of existing pre-trained models during inference time by updating the value of the predicted logit for each class that was obtained at the last layer of the neural network based on the predicted logit of its corresponding parent class. The aim is to calculate the conditional predicted probability for each class $k $ and instance $i $, taking into account the predicted probability of the parent class. We can formalize this by defining a new predicted probability for the $k$-th class $(c_k) $ and instance $i $ as follows.
\begin{equation}
    \widehat{p}_k^{(i)} = \frac{1}{ 1 + \exp \left(-\left(q_k^{(i)} + \alpha_k q_j^{(i)} \right)\right) }
    \label{eq:taxonomy.eq.1.pred.approach1}
\end{equation}
where $c_j = \Lambda (c_k)$ is the index of the parent class of the $k$-th class, and $\alpha_k $ is the hyperparameter that controls the influence of different parent class logits on child class logits.

When $\alpha_k=0 $, there is no influence from the parent class $c_j$ on the child class $c_k$.  By carefully selecting appropriate hyperparameter values, this transfer learning technique can be employed to effectively adjust the predicted probabilities of each class, considering the hierarchical relationship between classes, and potentially improving classification accuracy.

\subsubsection{Parameter Selection and Tuning}
The selection of appropriate hyperparameters is crucial for the effectiveness of the proposed transfer learning technique. In this study, we employ a systematic approach to tune the hyperparameter vector ${\{\alpha_k \}}_{k=1}^K$, which controls the dependency between the predicted probabilities of the child and parent classes. We utilize a grid search method to determine the optimal values for these hyperparameters. The search space for the hyperparameters is defined based on preliminary experiments and domain knowledge, ensuring a balance between model complexity and predictive performance. The proposed logit-based technique for optimizing the hyperparameters $\{\alpha_k\}_{k=1}^K$ is described in Algorithm~\ref{alg:logit_based_technique}.
\begin{enumerate}
    \item The algorithm takes as input the training, validation, and test datasets ($\mathbb{D}^{\text{train}}, \mathbb{D}^{\text{valid}}, \mathbb{D}^{\text{test}}$) obtained from the data preprocessing and dataset creation step (Algorithm~\ref{alg:taxonomy.dataset}), as well as the baseline model architecture $f(\cdot)$.
    \item The first step involves training the baseline model $f(\cdot)$ on the combined training and validation datasets ($\mathbb{D}^{\text{train}}$ and $\mathbb{D}^{\text{valid}}$). After training, the last sigmoid activation layer is dropped from the model, allowing access to the raw logit values.
    \item For each sample $(\mathbf{x}, \mathbf{y})$ in the training and test datasets, the logit vector $\mathbf{q}$ is computed by passing the input image $\mathbf{x}$ through the modified model $f(\cdot)$. The corresponding label vector $\mathbf{y}$ and logit vector $\mathbf{q}$ are then stored in the respective training or test set ($\mathbb{T}^{\text{train}}$ or $\mathbb{T}^{\text{test}}$).
    \item The algorithm then iterates over each pathology $c_k$ for $k \in \{1, \dots, K\}$. If the set of ancestor nodes $\Lambda(c_k)$ is not empty, an objective function $\mathcal{O}(\alpha_k)$ is defined as $1 - \text{AUC}(\{\hat{p}_k^{(i)}\}_{i=1}^N, \{y_k^{(i)}\}_{i=1}^N)$, where $\hat{p}_k^{(i)}$ and $y_k^{(i)}$ are the predicted probability and ground truth label for the $i$-th sample, respectively. The AUC (Area Under the ROC Curve) is used as the evaluation metric to assess the performance of the model for each pathology. The AUC metric was chosen for its robustness in assessing the quality of model predictions, irrespective of class imbalance, which is a common issue in medical image analysis.
    \item The optimal value of $\alpha_k$ is determined by minimizing the objective function $\mathcal{O}(\alpha_k)$ over the range $[-1, 1]$ using the Tree-structured Parzen Estimator (TPE) optimization algorithm. TPE is a Bayesian optimization technique that efficiently searches the hyperparameter space by constructing a surrogate model of the objective function and iteratively refining it based on the evaluated points.
    \item Finally, the algorithm returns the set of optimal hyperparameters $\{\alpha_k\}_{k=1}^K$ for each pathology. These hyperparameters are used to adjust the logit values and improve the model's performance on the hierarchical classification task.
\end{enumerate}

The logit-based technique provides a systematic approach to optimize the hyperparameters by leveraging the information contained in the logit values and the hierarchical structure of the pathologies. By minimizing the objective function based on the AUC metric, the algorithm aims to find the best values of $\alpha_k$ that enhance the model's ability to accurately predict the presence or absence of each pathology in the hierarchy.

\begin{SgAlgorithm}
    \caption{Data Preprocessing and Dataset Creation}\label{alg:taxonomy.dataset}
    \KwIn{$\mathbb{D}_{\text{NIH}}, \mathbb{D}_{\text{CheXpert}}, \mathbb{D}_{\text{PADCHEST}}$}
    \KwOut{$\mathbb{D}^{\text{train}}, \mathbb{D}^{\text{valid}}, \mathbb{D}^{\text{test}}$}
    $\mathbb{D} \gets \emptyset$\;
    \ForEach{$\mathbb{D}_i \in \{\mathbb{D}_{\text{NIH}}, \mathbb{D}_{\text{CheXpert}}, \mathbb{D}_{\text{PADCHEST}}\}$}{
        \ForEach{$(\mathbf{x}, \mathbf{y}) \in \mathbb{D}_i$}{
            Resize image $\mathbf{x}$ to $224 \times 224$ pixels\;
            Normalize pixel intensities of $\mathbf{x}$ to $[0, 1]$\;
            Standardize labels $\mathbf{y}$ to a common schema\;
            Extract 18 common pathologies from $\mathbf{y}$\;
            $\mathbb{D} \gets \mathbb{D} \cup \{(\mathbf{x}, \mathbf{y})\}$\;
        }
    }
    Randomly split $\mathbb{D}$ into $\mathbb{D}^{\text{train}}, \mathbb{D}^{\text{valid}}, \mathbb{D}^{\text{test}}$ with ratios 60\%, 10\%, 30\%\;
    \KwRet{$\mathbb{D}^{\text{train}}, \mathbb{D}^{\text{valid}}, \mathbb{D}^{\text{test}}$}
\end{SgAlgorithm}

\begin{SgAlgorithm}
    \caption{Logit-Based Technique}\label{alg:logit_based_technique}
    \KwIn{$\mathbb{D}^{\text{train}}, \mathbb{D}^{\text{valid}}, \mathbb{D}^{\text{test}}$, Baseline model architecture $f(\cdot)$}
    \KwOut{Optimal hyperparameters ${\alpha_k}_{k=1}^K$}
    Train $f(\cdot)$ on $\mathbb{D}^{\text{train}}$ and $\mathbb{D}^{\text{valid}}$\;
    Drop the last sigmoid activation layer\;
    \ForEach{$(\mathbf{x}, \mathbf{y}) \in \mathbb{D}^{\text{train}} \cup \mathbb{D}^{\text{test}}$}{
        Compute logit vector $\mathbf{q} = f(\mathbf{x})$\;
        Store $(\mathbf{y}, \mathbf{q})$ in $\mathbb{T}^{\text{train}}$ or $\mathbb{T}^{\text{test}}$ accordingly\;
    }
    \For{$k \in \{1, \dots, K\}$}{
        Initialize $\alpha_k = 0$\;
        \If{$\Lambda(c_k) \neq \varnothing$}{
            Define objective function $\mathcal{O}(\alpha_k) = 1 - \text{AUC}(\{\hat{p}_k^{(i)}\}_{i=1}^N, \{y_k^{(i)}\}_{i=1}^N)$\;
            $\alpha_k = \argmin_{\alpha_k \in [-1, 1]} \mathcal{O}(\alpha_k)$ using TPE\;
        }
    }
    \KwRet{$\{\alpha_k\}_{k=1}^K$}
\end{SgAlgorithm}

\subsection{Approach 2: Conditional Loss}\label{subsec:taxonomy.method.approach2}
In a second approach, we propose a similar concept to the approach discussed in Section~\ref{subsec:taxonomy.method.approach1}; however, rather than directly updating the predicted probability of each class, we instead update the loss value of each class based on the loss values of its parent classes. The novel hierarchical regularization term introduced in the loss-based approach directly incorporates label relationships into the model training process. This regularization term is designed to penalize misalignment between predictions and the label taxonomy, providing a new way to leverage hierarchical information during training. The utilization of the loss function approach can prove advantageous in certain scenarios, particularly in the context of multi-label classification tasks that involve hierarchical relationships, as it offers numerous benefits over existing methods:
\begin{itemize}
\item \textbf{Novel error minimization strategy:} The proposed approach introduces a new way to quantify and minimize the divergence between model predictions and ground truth labels by integrating parent class loss values into child class loss calculations. This novel hierarchical regularization term aims to improve prediction accuracy for both parent and child classes by penalizing misalignments throughout the hierarchy.
\item \textbf{Improved gradient propagation:} The loss-based approach incorporates parent class loss values into the calculation of child class losses, which can enhance gradient propagation between parent and child classes during backpropagation. This novel approach may lead to more effective learning of hierarchical associations and faster convergence compared to traditional methods.
\item \textbf{Increased robustness to label noise:} The proposed technique introduces a new way to handle inconsistencies or noise in ground truth labels by leveraging parent class loss values to compute child class losses. This approach enhances the consistency of the hierarchy by penalizing deviations from expected parent-child associations, potentially improving the model's resilience to label inaccuracies in the dataset.
\item \textbf{Improved interpretability:} By using loss values instead of predicted probabilities, the proposed approach enables a more straightforward understanding of the model's ability to capture hierarchical relationships among classes. The novel hierarchical regularization term ensures that high parent class loss values influence their corresponding child class losses, underscoring the importance of improving the underlying model architecture and parameters to better represent these hierarchical associations.
\end{itemize}

\subsubsection{Formulation of the Proposed Technique}
In multi-label classification problems, where each sample may belong to multiple classes, it is often necessary to combine the loss values for all classes to effectively train the model. Various methods can be employed to achieve this, depending on the specific problem. A common approach is to calculate the average loss across all classes for each sample by summing the losses for each class of a given sample and dividing the sum by the total number of classes to which the sample belongs. This method is effective when all classes are independent, of equal importance, and warrant equal weight in the total loss calculation. For example, in the case of cross-entropy loss, we have
\begin{equation}
    l_k \textcolor{gray}{\; = \; \mathcal{L} \left(y_k^{(i)},p_k^{(i)}\right)} \; = \; -\left(y_k^{(i)}\log(p_k^{(i)}) + (1 - y_k^{(i)})\log(1 - p_k^{(i)})\right)
    \label{eq:taxonomy.eq.2.loss}
\end{equation}
\begin{equation}
    \text{Loss}(\theta) = \sum_{i=1}^{N}\sum_{k=1}^{K}l_k
    \label{eq:taxonomy.eq.3.totalloss}
\end{equation}

In this formulation, the objective is to minimize the loss function with respect to the model parameters $\theta $, resulting in an optimal set of parameters that produce accurate predictions for multi-label classification tasks. However, class independence and equal importance between different classes cannot always be assumed. Inclusion of a hierarchical penalty or regularization term in the loss function is one way to push the loss function to take the taxonomy into account when optimizing the model hyperparameters (weights and biases).

Let's denote $\beta_k$ as the regularization term that penalizes the loss for class $c_k$ for each instance $i $ in which there is a low probability that it also belongs to parent class $c_j$. This can be represented mathematically by adding a hierarchical penalty term $H(c_k \vert c_j)$ for the class $c_k$ with respect to its corresponding parent class $c_j$ as follows:
\begin{equation}
    \widehat{l}_{k}^{(i)} = l_{k}^{(i)}+\beta_k H \left(c_k \vert c_j \right)
    \label{eq:taxonomy.eq.3.newloss}
\end{equation}
where $c_j=\Lambda(c_k)$, and $\beta_k $ is the hyperparameter that balances the contributions of class $k$'s own loss value and its parent class's loss values.

There are multiple ways to define the hierarchical penalty. For example, we can define it as the loss value of the parent class $l_j=\mathcal{L} \left(y_j^{(i)},p_j^{(i)}\right) $ as follows:
\begin{equation}
    H(k \vert j)=\mathcal{L} \left(y_j^{(i)},p_j^{(i)}\right)
    \label{eq:taxonomy.eq.4.hierarchical_penalty1}
\end{equation}

Another approach to incorporating the interdependence between different classes into the loss function is to apply the loss function $\mathcal{L} $ to the true label of the parent class and the predicted probability of the child class as follows.
\begin{equation}
    H(k\vert j)  = \mathcal{L} \left(y_j^{(i)},p_k^{(i)}\right)
    \label{eq:taxonomy.eq.5.hierarchical_penalty2}
\end{equation}

The penalization term in Equations~(\ref{eq:taxonomy.eq.4.hierarchical_penalty1}) and~(\ref{eq:taxonomy.eq.5.hierarchical_penalty2}) encourages the model to correctly predict the corresponding parent class when predicting the child class, hence ensuring that the predicted labels  align well with the hierarchical structure. The aforementioned approach, assumes a linear relationship  between the child and parent losses. However, this may not always accurately capture the relationship between the parent-child classes, as the relationship may not necessarily be linear.

The approach of multiplying losses introduces a greater adaptability in the representation of relationships between parent and child classes, as it can encapsulate both linear and potentially complex interrelations. Under the constraints of our problem --- where the absence of a parent class guarantees the absence of its child class --- both parent and child loss values would simultaneously increase or decrease (if the parent class is absent). In such a scenario, their summation or product would correspondingly escalate or diminish, thus demonstrating a linear relationship.

However, the complexity arises when we consider the scenario where the parent's loss value is significantly low in comparison to the child's loss. Here, a simple additive model might undervalue the parent's loss impact, as adding a small parent loss value to a considerably larger child loss value might not significantly alter the new updated loss for that child class. On the contrary, a multiplicative model amplifies the influence of each parent loss on the total, even if the parent's loss is relatively small. By defining the new loss for child classes in such way that their updated loss values are proportional to their corresponding parent's losses, we may enhance the hierarchical relationships' portrayal. To define such a loss value measurement scheme, we can modify the loss measurements presented in Equations~(\ref{eq:taxonomy.eq.4.hierarchical_penalty1}) and~(\ref{eq:taxonomy.eq.5.hierarchical_penalty2}) to be based on the multiplication of losses rather than their addition.
\begin{equation}
    \label{eq:taxonomy.eq.7.newloss}
    \widehat{l}_k^{(i)} = l_k^{(i)} H( k \vert j)
\end{equation}
where the hierarchical penalty term is

\begin{equation}
    \label{eq:taxonomy.eq.8.hierarchical_penalty.loss}
    H(k \vert j) =
    \left\{ \begin{array}{ll}
    1 & \text{otherwise.}
    \\
    \alpha_k l_j^{(i)} + \beta_k & c_j \text{ is parent of } c_k
    \end{array} \right.
\end{equation}
where $c_j$ is the parent class of the child class $c_k$, and $l_j$ is the parent loss value for instance $i$. \\

In Equation~(\ref{eq:taxonomy.eq.7.newloss}), $\widehat{l}_k^{(i)}$ represents the new loss value that we calculate by multiplying the original loss value $l_k^{(i)}$ for child class $k$ and instance $i$ with the hierarchical penalty term $H(k \vert j)$ which is calculated based on the parent class $j$. The hierarchical penalty term $H(k \vert j)$, defined in Equation~(\ref{eq:taxonomy.eq.8.hierarchical_penalty.loss}), adjusts based on the hierarchical relationships between classes. The terms $\alpha _k$ and $\beta_k$ are parameters that can be adjusted to control the degree of influence the hierarchical relationships have on the learning process.

The parameter $\alpha _k$ directly scales the parent's loss $l_j^{(i)}$. If $\alpha _k$ is increased, the penalty term becomes larger, and thus the updated loss $\widehat{l}_k^{(i)}$ becomes more sensitive to the parent's loss. This, in effect, increases the degree of influence that hierarchical information has on the learning process. The parameter $\beta_k$ serves as a baseline or offset. If $\beta_k$ is increased, the penalty term increases irrespective of the parent's loss value. This means that even if the parent's loss is low, the updated loss $\widehat{l}_k^{(i)}$ can still be high, thus maintaining the influence of hierarchical information in the learning process. However, if $\beta_k$ is set too high, it may lead to an overemphasis on hierarchy, possibly at the expense of other important learning elements. The regulation of parameters $\alpha_k$ and $\beta_k$ allow us to balance the degree to which hierarchical information influences the learning process, thus improving the reflection of the hierarchical structure in the model outputs, while remaining flexible to diverse learning scenarios.

\subsubsection{Parameter Selection and Tuning}
The selection of appropriate hyperparameters is crucial for the effectiveness of the proposed transfer learning technique. In this study, we employ a systematic approach to tune the two hyperparameter vectors $\alpha_k $, and $\beta_k$, which controls the dependency between the predicted probabilities of the child and parent classes. We utilize a grid search method to determine the optimal values for these hyperparameters. The search space for the hyperparameters is defined based on preliminary experiments and domain knowledge, ensuring a balance between model complexity and predictive performance.

The loss-based technique for optimizing the hyperparameters $\{\alpha_k\}_{k=1}^K$ and $\{\beta_k\}_{k=1}^K$ is described in Algorithm~\ref{alg:loss_based_technique}. The algorithm takes as input the training, validation, and test datasets ($\mathbb{D}^{\text{train}}, \mathbb{D}^{\text{valid}}, \mathbb{D}^{\text{test}}$) obtained from the data preprocessing and dataset creation step (Algorithm~\ref{alg:taxonomy.dataset}), the baseline model architecture $f(\cdot)$, and the baseline loss function $\mathcal{L}(\cdot)$.
\begin{enumerate}
    \item Similar to the logit-based technique, the loss-based technique begins by training the baseline model $f(\cdot)$ on the combined training and validation datasets ($\mathbb{D}^{\text{train}}$ and $\mathbb{D}^{\text{valid}}$), and then dropping the last sigmoid activation layer.
    \item For each sample $(\mathbf{x}, \mathbf{y})$ in the training and test datasets, the predicted probability vector $\mathbf{p}$ is computed by passing the input image $\mathbf{x}$ through the modified model $f(\cdot)$. The corresponding loss vector $\mathbf{l}$ is then calculated using the baseline loss function $\mathcal{L}(\mathbf{y}, \mathbf{p})$. The label vector $\mathbf{y}$ and loss vector $\mathbf{l}$ are stored in the respective training or test set ($\mathbb{T}^{\text{train}}$ or $\mathbb{T}^{\text{test}}$).
    \item The algorithm iterates over each pathology $c_k$ for $k \in \{1, \dots, K\}$. If the set of ancestor nodes $\Lambda(c_k)$ is not empty, an objective function $\mathcal{O}(\alpha_k, \beta_k)$ is defined as $1 - \text{AUC}(\{\hat{p}_k^{(i)}\}_{i=1}^N, \{y_k^{(i)}\}_{i=1}^N)$, where $\hat{p}_k^{(i)}$ and $y_k^{(i)}$ are the predicted probability and ground truth label for the $i$-th sample, respectively. The AUC metric is used to evaluate the performance of the model for each pathology.
    \item The optimal values of $\alpha_k$ and $\beta_k$ are determined by minimizing the objective function $\mathcal{O}(\alpha_k, \beta_k)$ over the range $[-1, 1] \times [-4, 4]$ using the Tree-structured Parzen Estimator (TPE) optimization algorithm. TPE efficiently searches the hyperparameter space to find the best combination of $\alpha_k$ and $\beta_k$ that minimizes the objective function.
    \item The algorithm returns the optimal hyperparameters $\{\alpha_k\}_{k=1}^K$ and $\{\beta_k\}_{k=1}^K$ for each pathology. These hyperparameters are used in the subsequent step of updating the predicted probabilities.
\end{enumerate}

Algorithm~\ref{alg:updating_probabilities} describes the process of updating the predicted probabilities using the optimal hyperparameters.
\begin{enumerate}
    \item For each sample $(\mathbf{x}, \mathbf{y})$ in the test set $\mathbb{D}^{\text{test}}$, the predicted probability vector $\mathbf{p}$ and the corresponding loss vector $\mathbf{l}$ are computed.
    \item For each pathology $c_k$, the ancestor node $c_j$ is determined using the function $\Lambda(c_k)$. The modified loss $\hat{l}_k$ is calculated as $l_k (\alpha_k l_j + \beta_k)$, where $l_k$ and $l_j$ are the losses for pathologies $c_k$ and $c_j$, respectively.
    \item The predicted probability $\hat{p}_k$ is updated based on a condition: if both $p_k$ and $p_j$ exceed their respective thresholds $\theta_k$ and $\theta_j$, then $\hat{p}_k$ is set to $\exp(-\hat{l}_k)$; otherwise, $\hat{p}_k$ is set to $1 - \exp(-\hat{l}_k)$. This condition ensures that the predicted probabilities are adjusted based on the presence or absence of the pathology and its ancestor node.
    \item Finally, the algorithm returns the updated predicted probabilities $\{\hat{p}_k^{(i)}\}_{k=1,i=1}^{K,N}$ for all pathologies and samples in the test set.
\end{enumerate}

The loss-based technique leverages the hierarchical structure of the pathologies and the loss values to optimize the hyperparameters and update the predicted probabilities. By minimizing the objective function based on the AUC metric and incorporating the modified loss, the algorithm aims to improve the model's performance on the hierarchical classification task.

\begin{SgAlgorithm}
    \caption{Loss-Based Technique}\label{alg:loss_based_technique}
    \KwIn{
        $\mathbb{D}^{\text{train}}, \mathbb{D}^{\text{valid}}, \mathbb{D}^{\text{test}}$ from Algorithm~\ref{alg:taxonomy.dataset}, Baseline model architecture $f(\cdot)$, Baseline loss function $\mathcal{L}(\cdot)$
        }
    \KwOut{
        Optimal hyperparameters ${\alpha_k}_{k=1}^K, {\beta_k}_{k=1}^K$
        }
    Train $f(\cdot)$ on $\mathbb{D}^{\text{train}}$ and $\mathbb{D}^{\text{valid}}$\;
    Drop the last sigmoid activation layer\;
    \ForEach{
        $(\mathbf{x}, \mathbf{y}) \in \mathbb{D}^{\text{train}} \cup \mathbb{D}^{\text{test}}$
    }{
        Compute $\mathbf{p} = f(\mathbf{x})$\;
        Compute loss $\mathbf{l} = \mathcal{L}(\mathbf{y}, \mathbf{p})$\;
        Store $(\mathbf{y}, \mathbf{l})$ in $\mathbb{T}^{\text{train}}$ or $\mathbb{T}^{\text{test}}$ accordingly\;
    }
    \For{$k \in \{1, \dots, K\}$}{
        Initialize $\alpha_k = 0, \beta_k = 1$\;
        \If{
            $\Lambda(c_k) \neq \varnothing$
        }{
            Define objective function $\mathcal{O}(\alpha_k, \beta_k) = 1 - \text{AUC}(\{\hat{p}_k^{(i)}\}_{i=1}^N, \{y_k^{(i)}\}_{i=1}^N)$\;
            Find $\alpha_k, \beta_k$ that minimize $\mathcal{O}(\alpha_k, \beta_k)$ over $[-1, 1] \times [-4, 4]$ using TPE\;
        }
    }
    \KwRet{
        $\{\alpha_k\}_{k=1}^K, \{\beta_k\}_{k=1}^K$
        }
\end{SgAlgorithm}

\begin{SgAlgorithm}
\caption{Updating Predicted Probabilities}\label{alg:updating_probabilities}
\KwIn{
    Test set $\mathbb{D}^{\text{test}}$, Trained model $f(\cdot)$, Optimal hyperparameters ${\alpha_k}_{k=1}^K, {\beta_k}_{k=1}^K$
    }
\KwOut{
    Updated predicted probabilities $\{\hat{p}_k^{(i)}\}_{k=1,i=1}^{K,N}$
    }
\ForEach{
    $(\mathbf{x}, \mathbf{y}) \in \mathbb{D}^{\text{test}}$
}{
    Compute $\mathbf{p} = f(\mathbf{x})$\;
    Compute loss $\mathbf{l} = \mathcal{L}(\mathbf{y}, \mathbf{p})$\;
    \ForEach{
        $k \in \{1, \dots, K\}$
    }{
        Determine $j = \Lambda(c_k)$\;
        Calculate modified loss $\hat{l}_k = l_k (\alpha_k l_j + \beta_k)$\;
        Update $\hat{p}_k$ based on condition: if $p_k > \theta_k \; \text{and} \; p_j > \theta_j$ then $\hat{p}_k = \exp(-\hat{l}_k)$ else $\hat{p}_k = 1 - \exp(-\hat{l}_k)$\;
    }
}
\KwRet{
    $\{\hat{p}_k^{(i)}\}_{k=1,i=1}^{K,N}$
    }
\end{SgAlgorithm}

\subsection{Updating Loss Values and Predicted Probabilities}\label{subsec:updating-loss-values-and-predicted-probabilities}
In the previous section, we introduced a taxonomy-based loss function with the goal of improving the classification accuracy of multi-class problems. However, one of the main advantages of our proposed technique is that it enables efficient utilization of pre-trained models and leverages the existing knowledge, thus reducing the computational cost and training time associated with re-optimization. In this section, we illustrate how both of our proposed approaches can be seamlessly integrated into an existing classification framework without the necessity to re-run the optimization phase of the classifier (e.g., DenseNet121). This can be achieved by focusing on updating the loss values (approach 2 shown in Section~\ref{subsec:taxonomy.method.approach2}) and predicted probabilities (approach 1 shown in Section~\ref{subsec:taxonomy.method.approach1}) to incorporate the hierarchical relationships present in the taxonomy structure.
During a training phase of a classifier (e.g., DenseNet121), an optimization algorithm such as gradient descent is used to determine the predicted probabilities that minimize the loss across the entire dataset. However, this approach is only valid during the training phase and only shows the predicted probability with respect to the original loss values measured by the classifier.
In the following, we show how to calculate the updated predicted probabilities from their updated loss values obtained from Equation~(\ref{eq:taxonomy.eq.7.newloss}) without re-doing the optimization process. Let us assume that binary cross entropy is used for the choice of the loss function $\mathcal{L}(\cdot) $. Let us denote $\widehat{q}_k^{(i)}, \widehat{p}_k^{(i)} $ as the updated values for logit and predicted probability of class $k $ and instance $i $ after applying the proposed technique. As previously discussed, to calculate the predicted probabilities, we need to pass the logits ${\widehat q}_k^{(i)} $ into a sigmoid function:
\begin{equation}
    \label{eq:taxonomy.eq.9.sigmoid}
    \widehat{p}_k^{(i)}=\text{sigmoid}\left(\widehat{q}_k^{(i)}\right)=\frac1{1+\exp\left(-\widehat{q}_k^{(i)}\right)}
\end{equation}

The sigmoid activation function maps any value to a number ranging from zero to one. The gradient of the sigmoid function (shown below) provides the direction in which the predicted probability must be updated.

\begin{align}
    \label{eq:taxonomy.eq.10.sigmoidprime}
    \frac{\partial{\text{sigmoid}}}{\partial{\widehat{q}_k^{(i)}}}
    & = \textcolor{gray}{\text{sigmoid}\left(\widehat{q}_k^{(i)}\right)\left(1-\text{sigmoid}\left(\widehat{q}_k^{(i)}\right)\right)}
    \\
    & = \widehat{p}_k^{(i)}\left(1-\widehat{p}_k^{(i)}\right)
\end{align}

The loss gradient gives us the direction in which the predicted probability needs to be updated to minimize the loss. The gradient of the binary cross-entropy loss is calculated as follows:
\begin{equation}
    \label{eq:taxonomy.eq.11.lossgradient}
    \frac{\partial \mathcal{L} \left( \widehat{p}_k^{(i)},\;y_k^{(i)}\right)}{\partial \widehat{p}_k^{(i)} }=\frac{y_k^{(i)}}{\widehat{p}_k^{(i)}}-\frac{1-y_k^{(i)}}{1-\widehat{p}_k^{(i)}}
\end{equation}
where $y_k^{(i)}\; $and $\widehat{p}_k^{(i)}\; $ are the true label and predicted probability, respectively, for instance $i $ and class $k $.

We now show how we can use the predicted probability, the gradient loss shown in Equation~(\ref{eq:taxonomy.eq.11.lossgradient}) and the derivative of the sigmoid function shown in Equation~(\ref{eq:taxonomy.eq.10.sigmoidprime}) to calculate the updated predicted probability as follows:
\begin{align}
    \label{eq:taxonomy.eq.12.newpredelement}
    \frac{\partial \mathcal{L}\left(p_k^{(i)},\; y_k^{(i)}\right)}{\partial \widehat{p}_k^{(i)} }\; \frac{\partial{\text{sigmoid}}}{\partial{\widehat{q}_k^{(i)}}}
    & \; = \; \textcolor{gray}{\left(\frac{y_k^{(i)}}{\widehat{p}_k^{(i)}}-\frac{1-y_k^{(i)}}{1-\widehat{p}_k^{(i)}}\right)\widehat{p}_k^{(i)}\left(1-\widehat{p}_k^{(i)}\right) }
    \\
    & \; = \; y_k^{(i)}-\widehat{p}_k^{(i)}
\end{align}

Hence, we can conclude that
\begin{equation}
    \label{eq:taxonomy.eq.13.newpred}
    \begin{array}{@{}l}
    \hat{p}_{k}^{(i)} = \left\{
        \begin{array}{ll}
           1-\, \frac{\partial \mathcal{L}\left(p_k^{(i)},\;y_k^{(i)}\right)}{\partial \widehat{p}_k^{(i)} }\;{ \frac{\partial{\text{sigmoid}}}{\partial{\widehat{q}_k^{(i)}}}}
            &
            y_k^{(i)}=1
            \\
            -\,\frac{\partial \mathcal{L}\left(p_k^{(i)},\;y_k^{(i)}\right)}{\partial \widehat{p}_k^{(i)} }\; {\frac{\partial{\text{sigmoid}}}{\partial{\widehat{q}_k^{(i)}}}}
            &
            \text{otherwise.}
        \end{array}\right.
    \end{array}
\end{equation}

We would like to modify this equation so that it does not directly depend on the true value and instead rely on the gradient loss. If we simplify the loss gradient shown in Equation~(\ref{eq:taxonomy.eq.11.lossgradient}) we obtain the following:
\begin{align}
    \label{eq:taxonomy.eq.14.newlossgradient}
    \frac{\partial \mathcal{L}(\widehat{p}_k^{(i)}, y_k^{(i)})}{\partial \widehat{p}_k^{(i)}}
    & \; = \; \textcolor{gray}{\frac{y_k^{(i)}}{\widehat{p}_k^{(i)}} - \frac{1 - y_k^{(i)}}{1 - \widehat{p}_k^{(i)}} }
    \\
    & \; = \; \frac{y_k^{(i)} - \widehat{p}_k^{(i)}}{\widehat{p}_k^{(i)}{\left(1 - \widehat{p}_k^{(i)}\right)}}
\end{align}

In this equation, we see that when the true label is positive $\left(y_k^{(i)}=1\right) $, the loss gradient can only be 0 or a positive number. Similarly, when zero $\left(y_k^{(i)}=0\right) $, the loss gradient can only take the value 0 or a negative number. Thus, we can modify Equation~(\ref{eq:taxonomy.eq.13.newpred})  as follows:
\begin{equation}
    \label{eq:taxonomy.eq.15.newpred}
    \widehat{p}_k^{(i)} =
    \begin{cases}
        -\, \frac{\partial \mathcal{L}(\widehat{p}_k^{(i)}, y_k^{(i)})}{\partial {\widehat p}_k^{(i)}} \, \frac{\partial{\text{sigmoid}}}{\partial{\widehat{q}_k^{(i)}}} + 1
        &
        \text{if} \quad \frac{\partial \mathcal{L}(\widehat{p}_k^{(i)}, y_k^{(i)})}{\partial {\widehat p}_k^{(i)}} \geq 0
        \\
        -\, \frac{\partial \mathcal{L}(\widehat{p}_k^{(i)}, y_k^{(i)})}{\partial {\widehat p}_k^{(i)}} \, \frac{\partial{\text{sigmoid}}}{\partial{\widehat{q}_k^{(i)}}}
        &
        \text{otherwise.}
    \end{cases}
\end{equation}

Finally, Equation~(\ref{eq:taxonomy.eq.15.newpred}) can be simplified as follows:
\begin{equation}
    \label{eq:taxonomy.eq.16.newpred}
    \widehat{p}_k^{(i)} =
    \begin{cases}
        \, \exp(-\widehat{l}_k^{(i)})
        &
        \text{if} \quad \frac{\partial \mathcal{L}(\widehat{p}_k^{(i)}, y_k^{(i)})}{\partial {\widehat p}_k^{(i)}} \geq 0
        \\
        \, 1 - \exp(-\widehat{l}_k^{(i)})
        &
        \text{otherwise}
    \end{cases}
\end{equation}
where, ${\widehat l}_k^{(i)} $ is the updated loss for class $k $ and instance $i $.

Alternatively, we can substitute condition factor in Eq~\ref{eq:taxonomy.eq.16.newpred} to have.
\begin{equation}
    \label{eq:taxonomy.loss.newpred_based_on_loss}
    \widehat{p}_k^{(i)} =
    \begin{cases}
        \, \exp(-\widehat{l}_k^{(i)})
        &
        \text{if} \quad y_{k}^{(i)}=1
        \\
        \, 1 - \exp(-\widehat{l}_k^{(i)})
        &
        \text{otherwise}
    \end{cases}
\end{equation}

In order to utilize this during inference time, we can substitute the $y_{k}^{(i)}=1$ condition with the predicted presence of both child and parent class.
\begin{equation}
    \label{eq:taxonomy.loss.newpred_based_on_loss_inference}
    \widehat{p}_k^{(i)} =
    \begin{cases}
        \, \exp(-\widehat{l}_k^{(i)})
        &
        \text{if} \quad (p_k^{(i)} > \theta_k) \; \& \; (p_j^{(i)} > \theta_j)
        \\
        \, 1 - \exp(-\widehat{l}_k^{(i)})
        &
        \text{otherwise}
    \end{cases}
\end{equation}
where $\theta_k$, and $\theta_j$ are binarization threshold values for child and parent classes respectively.

\subsubsection{Updated Predicted Probabilities with Respect to Original Values}
The following demonstrates Equation~(\ref{eq:taxonomy.eq.16.newpred}) based on predicted probability to demonstrate its similarity to Equation~(\ref{eq:taxonomy.eq.1.pred.approach1}) in Approach 1 (Section~\ref{subsec:taxonomy.method.approach1}). From Equation~(\ref{eq:taxonomy.eq.8.hierarchical_penalty.loss}) we have $\hat{l}_k^{(i)}=l_k^{(i)}\left(\alpha_k\;l_j^{(i)}+\beta_k\right) $. By substituting that into $\exp{\left(-\widehat{l}_{k}^{(i)}\right)}, \text{for } y_{k}^{(i)}=1 $ we obtain:
\begin{align}
    \label{eq:taxonomy.eq.17}
    \exp{\left(-{\widehat l}_k^{(i)}\right)}
    & \; = \; \textcolor{gray}{\exp{\left(-l_k^{(i)}\left(\alpha_k \; l_j^{(i)}+\beta_k\right)\right)}}
    \\
    & \; = \; {\left(p_k^{(i)}\right)}^{-\alpha_k{\log{\left(p_j^{(i)}\right)}}+\beta_k}
\end{align}

Furthermore, $1-\exp{\left(-{\widehat l}_k^{(i)}\right)},\text{for}\; y_k^{(i)}=0 $ is as follows:
\begin{align}
    \label{eq:taxonomy.eq.18}
    1-\exp{\left(-{\widehat l}_k^{(i)}\right)}
    & \; = \; \textcolor{gray}{1-\exp{\left(-l_k^{(i)}\left(\alpha_k\;l_j^{(i)}+\beta_k\right)\right)} \nonumber}
    \\
    & \; = \; {1-\left(1-p_k^{(i)}\right)}^{-\alpha_k{\log{\left(1-p_j^{(i)}\right)}}+\beta_k}
\end{align}

By substituting Equations~(\ref{eq:taxonomy.eq.17}) and~(\ref{eq:taxonomy.eq.18})  into Equation~(\ref{eq:taxonomy.eq.16.newpred})  we obtain
\begin{equation}
    \label{eq:taxonomy.eq.19.newpred}
    \widehat{p}_k^{(i)} =
    \begin{cases}
        \, {\left( p_k^{(i)} \right)}^{-\alpha_k \log(p_j^{(i)}) + \beta_k}
        &
        \text{if} \quad y_k^{(i)} = 1
        \\
        \, 1 - {\left( 1 - p_k^{(i)} \right)}^{-\alpha_k \log{\left( 1 - p_j^{(i)} \right)} + \beta_k}
        &
        \text{otherwise.}
    \end{cases}
\end{equation}

In order to make this work on new test instances, we can substitute the $y_{k}^{(i)}=1$ condition with the predicted presence of both child and parent class.
\begin{equation}
    \label{eq:taxonomy.eq.19.newpred_wo_groundtruth}
    \widehat{p}_k^{(i)} =
    \begin{cases}
        \, {\left( p_k^{(i)} \right)}^{-\alpha_k \log(p_j^{(i)}) + \beta_k}
        &
        \text{if} \quad (p_k^{(i)} > 0.5) \; \& \; (p_j^{(i)} > 0.5)
        \\
        \, 1 - {\left( 1 - p_k^{(i)} \right)}^{-\alpha_k \log{\left( 1 - p_j^{(i)} \right)} + \beta_k}
        &
        \text{otherwise.}
    \end{cases}
\end{equation}

\subsection{Experimental Setup}
\subsubsection{Datasets}
Three diverse and publicly available datasets are used to evaluate the proposed hierarchical multi-label classification techniques: CheXpert~\cite{irvin_CheXpert_2019}, PADCHEST~\cite{bustos_Padchest_2020}, and NIH~\cite{wang_ChestXRay8_2017}. These datasets contain a diverse range of chest radiographic images covering various thoracic diseases, providing a comprehensive evaluation of the effectiveness of our method. The description of the three datasets are as follows.
\begin{itemize}
    \item  \textbf{CheXpert}~\cite{irvin_CheXpert_2019} is a large-scale dataset containing 224,316 chest radiographs of 65,240 patients, labeled with 14 radiographic findings.
    \item \textbf{PADCHEST}~\cite{bustos_Padchest_2020} consists of 160,000 chest radiographs of 67,000 patients, annotated with 174 radiographic findings. This dataset is highly diverse and includes a wide variety of thoracic diseases.
    \item \textbf{NIH}~\cite{wang_ChestXRay8_2017} includes 112,120 chest radiographs of 30,805 patients labeled with 14 categories of thoracic diseases.
\end{itemize}

\textit{Preprocessing: }
The primary algorithm for data preprocessing and dataset creation (Algorithm~\ref{alg:taxonomy.dataset}) was designed to integrate and standardize heterogeneous chest radiograph datasets, namely $\mathbb{D}{\text{NIH}}$, $\mathbb{D}{\text{CheXpert}}$, and $\mathbb{D}_{\text{PADCHEST}}$. The images were resized to a resolution of $224 \times 224$ pixels, with the pixel intensities normalized to a range of 0 and 1. Data augmentation techniques, such as rotation, translation, and horizontal flipping, were applied to increase the dataset's size and diversity, consequently enhancing the model's generalization capability. The choice of resizing images to $224 \times 224$ pixels ensures compatibility with prevalent deep learning models, balancing computational efficiency with the retention of critical diagnostic features. The normalization of pixel intensities to the range $[0, 1]$ is a standard practice to facilitate model convergence by providing a consistent scale for input data. Label standardization to a common schema and the extraction of 18 common pathologies were paramount for ensuring consistency across datasets, thereby enabling the comparative analysis of model performance.

The random split of the combined dataset into training, validation, and testing subsets with ratios of 60
\subsubsection{Model Optimization}
The DenseNet121~\cite{huang_Densely_2017} architecture and the pre-trained weights provided by Cohen~\cite{cohen_TorchXRayVision_2022} was used as the baseline model. The model was fine-tuned on a subset of CheXpert~\cite{irvin_CheXpert_2019}, NIH~\cite{wang_ChestXRay8_2017}, PADCHEST~\cite{bustos_Padchest_2020} for 18 thoracic diseases. A series of transformations were applied to all train images, including rotation of up to 45 degrees, translation of up to 15\%, and scaling up to 10\%. Binary cross entropy losses and Adam optimizer were used.

\subsubsection{Parallelization for multiple CPU cores:}
To effectively optimize the hyperparameters of our proposed taxonomy-based transfer learning methods, we utilize parallelization techniques that distribute the computational load across multiple CPU cores. By leveraging the power of parallel processing, we can drastically reduce the overall computation time and accelerate the optimization procedure, making the method more applicable to large-scale and high-dimensional datasets. Different parallelization libraries, such as joblib and Python multiprocessing, were employed to facilitate the implementation of parallelism, ensuring seamless integration with existing frameworks and offering a scalable and hardware-adaptable solution.

\subsubsection{Optimum Threshold Determination:}
Determining the optimal threshold is a crucial aspect of evaluating the performance of the proposed method, as it determines the point at which the predictions for multi-label classification tasks are translated into binary class labels. To determine the optimal threshold value, we used receiver operating characteristic (ROC) analysis, a common method for evaluating the performance of classification models. ROC analysis provides a comprehensive view of the model's performance at various threshold values, allowing us to determine the optimal point for balancing the true positive rate (sensitivity) and the false positive rate (specificity) (1-specificity). By plotting the ROC curve and calculating the area under the curve (AUC), we can quantitatively evaluate the discriminatory ability of the model and compare its performance at various threshold values. The optimal threshold is determined by locating the point on the ROC curve closest to the upper left corner, which represents the highest true positive rate and the lowest false positive rate. By incorporating ROC analysis and optimal threshold determination into our experimental design, we ensure that our results not only accurately reflect the performance of the model but also provide valuable insight into the practical applicability of our approach in real-world settings.

\subsubsection{Evaluation:}
To assess the performance of the proposed techniques in accurately classifying samples compared to a baseline model, several evaluation metrics were used. The metrics were selected based on their ability to provide a comprehensive assessment of the model's performance in terms of accuracy, precision, recall, and the ability to differentiate between true and false positives. The evaluated metrics are as follows.
\begin{itemize}
    \item \textbf{Accuracy} measures the proportion of correctly classified samples to the total number of samples.
    \item \textbf{F1-score} is the harmonic mean of precision and recall, providing a balanced assessment of the method's performance.
    \item \textbf{Area Under the Receiver Operating Characteristic Curve (AUROC)}: The ROC curve is a graphical representation of the diagnostic performance of a binary classifier system as its discrimination threshold is varied. The ROC curve is derived by plotting the true positive rate (TPR) versus the false positive rate (FPR) at different thresholds. The AUC provides a single scalar value representing the expected performance of the classifier. An AUC of 1 indicates that the classifier can distinguish perfectly between the two classes (e.g., ``positive'' and ``negative''), whereas an AUC of 0.5 indicates that the classifier is no better than random chance.
    \item \textbf{t-stat (t-statistic)} is a measurement of the magnitude of the difference relative to the variance in sample data. The t-value quantifies the statistical significance of the difference. It is used to test hypotheses regarding the mean or the difference between two means when the standard deviation of the population is unknown.
    \item \textbf{p-value}: In hypothesis testing, the p-value is a function used to determine the significance of the results. It represents the probability that test results were generated at random. If the p-value is small (typically 0.05), there is strong evidence that the null hypothesis should be rejected.
    \item \textbf{Cohen's Kappa} measures the concordance between two raters who classify items into mutually exclusive categories. Primarily, it is used to determine the degree of agreement between two raters. The Kappa score takes into account the possibility that the agreement occurred by chance. A Kappa score of 1 indicates perfect concordance between two raters. A Kappa score of less than 1 indicates less than perfect agreement, and a Kappa score of less than 0 indicates either no agreement or agreement that is worse than random.
    \item \textbf{BF10 (Bayes Factor)} rates the strength of the evidence in favor of one statistical model over another, given the available data. BF10 specifically contrasts the evidence supporting a null hypothesis (H0) with an alternative hypothesis (H1). The data are equally likely to be true under the null and alternative hypotheses, according to a BF10 value of 1. A BF10 value greater than 1 denotes support for H1, while a value lower than 1 denotes support for H0. In general, values between 1/3 and 3 are regarded as inconclusive, values above 3 as some evidence for H1, and values below 1/3 as evidence for H0.
    \item \textbf{Cohen's d} is a measure of effect size in the context of a t-test for the difference between two means. It can be calculated as the difference between two means divided by the data's standard deviation. Typically, small, medium, and large effect sizes are referred to as Cohen's d values of 0.2, 0.5, and 0.8, respectively. It is a common method of estimating the difference between two groups after adjusting for variance and sample size variations.
    \item \textbf{Power (Statistical Power)} is the likelihood that a test will correctly reject the null hypothesis when the alternative hypothesis is true (i.e., the test will not make a Type II error). Power is typically desired to be 0.8 or higher, meaning there is an 80\% or greater chance of discovering a true effect if it is present. Many variables, such as the effect size, sample size, significance level, and data variability, can have an impact on power. Calculating power can be used to determine the sample size required to detect an effect of a given size when designing a study.
\end{itemize}

\textit{Some limitations of these metrics are as follows.}
While accuracy is a useful metric for evaluating overall performance, it may not be the most appropriate metric for unbalanced datasets in which the number of samples in each class is significantly different. Similarly, F1-score may be biased towards the class with a larger sample size, and AUROC may not be appropriate for datasets with a high degree of class overlap. In addition, outliers or non-normal distributions may influence the t-statistic and p-value, whereas Cohen's Kappa may not be applicable to non-categorical data. BF10 may be affected by the selection of prior probabilities, and Cohen's d may not apply to non-parametric data. The choice of significance level and the data variability may have an effect on the power.
\section{Results}\label{sec:taxonomy.results}
\subsection{Taxonomy Structure}
\begin{figure}[H]
    \centering
    \includegraphics[width=\textwidth]{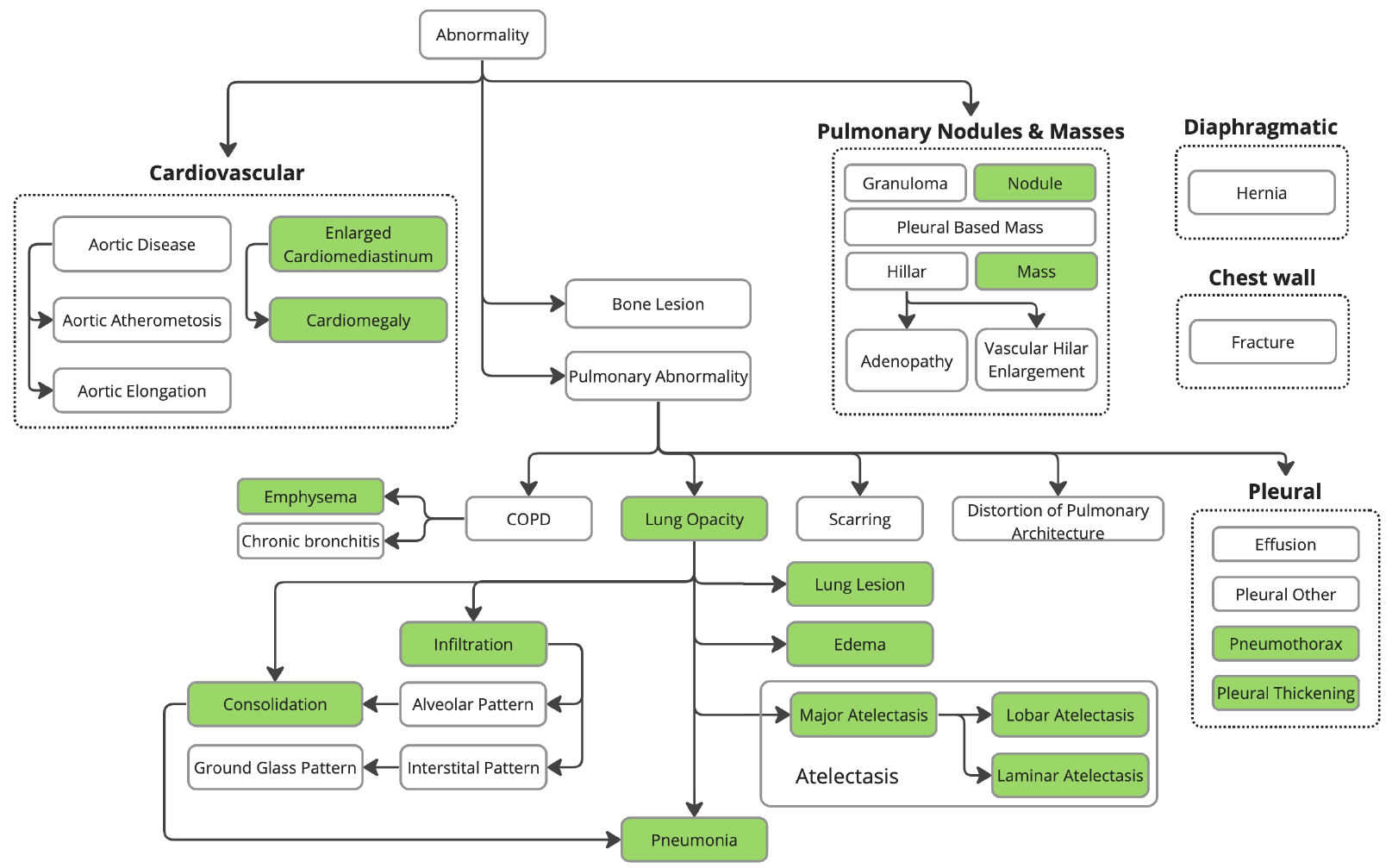}
    \caption[Taxonomy Structure of Lung Pathologies in Chest Radiographs]{Taxonomy structure of lung pathologies in chest radiographs.}\label{fig:taxonomy.fig.1.taxonomy_structure}
\end{figure}

In this study, we have developed a comprehensive taxonomy, as illustrated in Figure~\ref{fig:taxonomy.fig.1.taxonomy_structure}, which serves as a pivotal foundation for categorizing a wide array of lung pathologies identifiable in public chest X-ray datasets. This classification system, draws from the foundational work Irvin~\cite{irvin_CheXpert_2019}, Chen~\cite{chen_Deep_2020}, and Gohagan~\cite{gohagan_Prostate_2000}, encompassing prevalent disease manifestations identified in widely used datasets such as CheXpert~\cite{irvin_CheXpert_2019}, PADCHEST~\cite{bustos_Padchest_2020}, and NIH~\cite{wang_ChestXRay8_2017}. Our taxonomy provides a structured approach to categorize these disease manifestations and offers a framework to facilitate the comprehension and analysis of abnormalities in chest radiographs.

The genesis of this taxonomy lies in the need for a systematic approach to distinguish and interrelate various lung pathologies, addressing the limitations of existing classification methods. By presenting the proposed taxonomy structure in Figure~\ref{fig:taxonomy.fig.1.taxonomy_structure}, we highlight a hierarchical organization that spans a broad spectrum of pulmonary abnormalities, including, but not limited to, nodules, masses, pleural effusions, and pneumonia. The design of this taxonomy intricately maps out the relationships between different pathologies, showcasing how specific conditions are methodically categorized under broader disease entities. This hierarchical arrangement not only aids in capturing the multifaceted nature of lung diseases but also underpins the development of our innovative multi-label classification techniques. These techniques exploit the structured information embodied in the taxonomy, significantly enhancing both the accuracy and interpretability of diagnostic predictions.

Delving into the components of Figure~\ref{fig:taxonomy.fig.1.taxonomy_structure}, the taxonomy meticulously outlines various levels of disease classifications, from broad categories such as Cardiovascular, Pulmonary Nodules \& Masses, and other Pulmonary abnormalities, down to more specific conditions like Granuloma, Nodule, and Pleural Based Mass. Each category is further detailed with subcategories, for instance, Aortic Disease within Cardiovascular, branching into Aortic Atherometosis and Aortic Elongation. The directional arrows within the figure signify parent-child relationships, emphasizing the logical flow from general to specific disease states. This detailed component analysis serves not only as an educational tool for comprehending the intricate relationships between different lung pathologies but also as a strategic framework for applying hierarchical multi-label classification techniques. By carefully selecting pathologies marked in green based on, we underline the taxonomy's application relevance and its critical role in advancing chest radiograph analysis.
For the purpose of this study, the pathologies marked in green were carefully selected based on criteria such as label availability, cross-dataset presence and having a child-parent relation.

\subsection{Datasets}
The prevalence of different pathology labels across three distinct medical imaging datasets: CheXpert~\cite{irvin_CheXpert_2019}, PADCHEST~\cite{bustos_Padchest_2020}, and NIH~\cite{wang_ChestXRay8_2017} are examined. Table~\ref{tab:taxonomy.table.1.datasets.pathologies} provides an overview of each pathology label's prevalence across these datasets. To utilize the TorchXRayVision software~\cite{cohen_TorchXRayVision_2022}, the same 18 pathologies as their work, were chosen for model fine-tuning. For the purpose of assessing the proposed methodologies, particular emphasis is placed on pathologies that are recurrent across different datasets (appear in at least two of the three datasets) and are included in our formulated taxonomy. Furthermore, the cross-dataset presence of these pathologies enhances the generalizability of our study, as the developed models are validated on multiple independent datasets. These pathologies, highlighted in \colorbox{mygreen}{green} in the table, comprise \textbf{Atelectasis}, \textbf{Consolidation}, \textbf{Infiltration}, \textbf{Edema}, \textbf{Pneumonia}, \textbf{Cardiomegaly}, \textbf{Lung Lesion}, \textbf{Lung Opacity}, and \textbf{Enlarged Cardiomediastinum}. The pathologies that are not highlighted, i.e., those occurring in just one or none of the datasets or not included in our taxonomy, were not included in the final evaluation of this study. Their exclusion is mainly due to the lack of sufficient data for a robust comparison or their non-alignment with the taxonomy structure studied.
\begingroup
\setlength{\LTleft}{\fill}
\setlength{\LTright}{\fill}
\begin{longtable}{lccc}
    \caption[Representation of Pathologies Across Datasets]{Representation of pathologies across datasets}%
    \label{tab:taxonomy.table.1.datasets.pathologies}\\
    \cellcolor{table_title}{\textbf{Pathologies}} &
    \cellcolor{table_title}{\textbf{NIH}}         &
    \cellcolor{table_title}{\textbf{PADCHEST}}    &
    \cellcolor{table_title}{\textbf{CheX}}        \\
    \endfirsthead
    \caption[]{Representation of pathologies across datasets (continued)}\\
    \cellcolor{table_title}{\textbf{Pathologies}} &
    \cellcolor{table_title}{\textbf{NIH}}         &
    \cellcolor{table_title}{\textbf{PADCHEST}}    &
    \cellcolor{table_title}{\textbf{CheX}}        \\
    \endhead
    Air    Trapping      & & X & \\
    Aortic Atheromatosis & & X & \\
    Aortic Elongation    & & X & \\
    Aortic Enlargement   & &   & \\
    \cellcolor{table_row_highlight}\textbf{Atelectasis} &
    \cellcolor{table_row_highlight} X &
    \cellcolor{table_row_highlight} X &
    \cellcolor{table_row_highlight} X \\
    Bronchiectasis           & & X & \\
    Calcification            & &   & \\
    Calcified Granuloma      & &   & \\
    \cellcolor{table_row_highlight}\textbf{Cardiomegaly} &
    \cellcolor{table_row_highlight} X &
    \cellcolor{table_row_highlight} X &
    \cellcolor{table_row_highlight} X \\
    \cellcolor{table_row_highlight}\textbf{Consolidation} &
    \cellcolor{table_row_highlight} &
    \cellcolor{table_row_highlight} X &
    \cellcolor{table_row_highlight} X \\
    Costophrenic Angle Blunting &  & X &  \\
    \cellcolor{table_row_highlight}\textbf{Edema} &
    \cellcolor{table_row_highlight} X      &
    \cellcolor{table_row_highlight} X      &
    \cellcolor{table_row_highlight} X      \\
    \textbf{Effusion} & X & X & X \\
    \textbf{Emphysema} & X & X &  \\
    \cellcolor{table_row_highlight}\textbf{Enlarged Cardiomediastinum} &
    \cellcolor{table_row_highlight} &
    \cellcolor{table_row_highlight} &
    \cellcolor{table_row_highlight} X \\
    \textbf{Fibrosis}         & X & X &   \\
    Flattened Diaphragm       &   & X &   \\
    Fracture                  &   & X & X \\
    Granuloma                 &   & X &   \\
    Hemidiaphragm Elevation   &   & X &   \\
    \textbf{Hernia}           & X & X &   \\
    Hilar Enlargement         &   & X &   \\
    ILD                       &   &   &   \\
    \cellcolor{table_row_highlight}\textbf{Infiltration} &
    \cellcolor{table_row_highlight} X &
    \cellcolor{table_row_highlight} X &
    \cellcolor{table_row_highlight} \\
    \cellcolor{table_row_highlight}\textbf{Lung Lesion} &
    \cellcolor{table_row_highlight} &
    \cellcolor{table_row_highlight} &
    \cellcolor{table_row_highlight} X \\
    \cellcolor{table_row_highlight}\textbf{Lung Opacity} &
    \cellcolor{table_row_highlight} &
    \cellcolor{table_row_highlight} &
    \cellcolor{table_row_highlight} X \\
    \textbf{Mass}               & X & X &   \\
    Nodule/Mass                 &   &   &   \\
    \textbf{Nodule}             & X & X &   \\
    \textbf{Pleural Other}      &   &   & X \\
    \textbf{Pleural Thickening} & X & X &   \\
    \cellcolor{table_row_highlight}\textbf{Pneumonia} &
    \cellcolor{table_row_highlight} X         &
    \cellcolor{table_row_highlight} X         &
    \cellcolor{table_row_highlight} X         \\
    \textbf{Pneumothorax}      & X & X & X \\
    Pulmonar Fibrosis          &   &   &   \\
    Scoliosis                  &   & X &   \\
    Tuberculosis               &   & X &   \\
    Tube                       &   & X &   \\
\end{longtable}
\endgroup

The distribution of samples per pathology in each dataset is presented in Table~\ref{tab:taxonomy.table.2.datasets.ninstances}. Before applying the proposed technique, a series of preprocessing steps are performed on the ground truth label set. In the context of medical images containing multiple classes, it is a prevailing practice for the individual responsible for labeling to solely annotate the pathologies that are pertinent to their specific study requirements. Occasionally, there are situations wherein certain instances of data are classified as having specific child pathologies, but not their corresponding parent pathologies. In order to address the absence of labels for certain parent classes, which is crucial for the efficacy of the proposed techniques, we have modified the label value to signify the presence of classes with at least one child class as \textcolor{blue}{TRUE}, indicating the existence of the class in that particular instance. A preprocessing step is applied to classes that do not have corresponding labels in the original ground truth label set. In the context of this study, the Lung Opacity and Enlarged Cardiomediastinum classes are absent from the original ground truth label sets of the NIH and PADCHEST datasets (Table~\ref{tab:taxonomy.table.1.datasets.pathologies}). By revising the ground truth label set, we have identified several instances where the presence of the respective parent class can be inferred based on the presence of their respective child classes as shown in Table~\ref{tab:taxonomy.table.2.datasets.ninstances} (cells highlighted in green).

\begin{table}[H]
    \centering
    \caption[Sample Distribution Per Pathology in Evaluated Datasets (CheX, NIH, and PC)]{Sample distribution per pathology in evaluated datasets (CheX, NIH, and PC)}%
    \label{tab:taxonomy.table.2.datasets.ninstances}
    \begin{tabular}{lcccccc}
        \rowcolor[HTML]{79A8A4}
        \multicolumn{1}{c}{\cellcolor{table_title}{}} &
        \multicolumn{2}{c}{\cellcolor{table_title}{\textbf{CheXpert}}} &
        \multicolumn{2}{c}{\cellcolor{table_title}{\textbf{NIH}}} &
        \multicolumn{2}{c}{\cellcolor{table_title}{\textbf{PADCHEST}}} \\
        \rowcolor[HTML]{79A8A4}
        \multicolumn{1}{c}{\multirow{-2}{*}{\cellcolor{table_title}{\textbf{Pathologies\textbackslash{}Dataset}}}} & {PA} & {AP} & {PA} & {AP} & {PA} & {AP} \\
        Atelectasis        & 2460 & 11643 & 1557 & 1016 & 2419 & 232 \\
        Consolidation      & 1125 & 4956  & 384  & 253  & 475  & 77  \\
        Infiltration       & 0    & 0     & 3273 & 1131 & 4309 & 587 \\
        Pneumothorax       & 1060 & 4239  & 243  & 253  & 97   & 15  \\
        Edema              & 1330 & 15117 & 39   & 237  & 108  & 130 \\
        Emphysema          & 0    & 0     & 264  & 193  & 546  & 30  \\
        Fibrosis           & 0    & 0     & 556  & 61   & 341  & 8   \\
        Effusion           & 5206 & 19349 & 1269 & 654  & 1625 & 311 \\
        Pneumonia          & 992  & 2064  & 175  & 89   & 1910 & 211 \\
        Pleural\_Thickening& 0    & 0     & 745  & 145  & 2075 & 34  \\
        Cardiomegaly       & 2117 & 8284  & 729  & 203  & 5387 & 261 \\
        Nodule             & 0    & 0     & 1609 & 460  & 2190 & 95  \\
        Mass               & 0    & 0     & 1213 & 493  & 506  & 17  \\
        Hernia             & 0    & 0     & 81   & 13   & 988  & 38  \\
        Lung Lesion        & 1655 & 3110  & 0    & 0    & 0    & 0   \\
        Fracture           & 1115 & 3463  & 0    & 0    & 1662 & 69  \\
        Lung Opacity& 7006 & 28183 &
        \cellcolor{table_row_highlight}4917 &
        \cellcolor{table_row_highlight}2216 &
        \cellcolor{table_row_highlight}6947 &
        \cellcolor{table_row_highlight}861 \\
        Enlarged Cardiomediastinum & 1100 & 4577 &
        \cellcolor{table_row_highlight}729 &
        \cellcolor{table_row_highlight}203 &
        \cellcolor{table_row_highlight}5387 &
        \cellcolor{table_row_highlight}261 \\
        \rowcolor[HTML]{79A8A4}
        Total & 20543 & 53359 & 28868 & 9060 & 61692 & 2445
    \end{tabular}
\end{table}

\subsection{Techniques Evaluation}
The performance comparison of our proposed methods, namely logit-based and loss-based, with the ``baseline'' technique is illustrated in Figure~\ref{fig:taxonomy.fig.3.roc_curve_all_datasets}. This comparative analysis centers on nine distinct medical conditions associated with pulmonary and cardiovascular diseases within three datasets. These nine pathologies encompass two parent classes (\textbf{Lung Opacity}, and \textbf{Enlarged Cardiomediastinum}) and their respective child classes, as illustrated in Figure~\ref{fig:taxonomy.fig.1.taxonomy_structure}. Each subplot exhibits the receiver operating characteristic (ROC) curves for each methodology superimposed on one another, accompanied by their respective AUC (Area Under Curve) scores annotated. AUC (Area Under the Curve) scores are computed for each pathology class across all test samples in all studies datasets. We can see a notable improvement in AUC scores for all pathologies possessing parent classes. The aforementioned findings serve as compelling evidence for the effectiveness of the proposed methodologies, as they showcase their ability to improve the accuracy of classification in scenarios involving hierarchical class structures. AUC scores for two parent classes, ``Lung Opacity'' and ``Enlarged Cardiomediastinum'', remain unchanged as expected. The techniques proposed in this study are designed to exploit the hierarchical structure of classes, and therefore only bring about improvements where a class possesses a parent class.

\begin{table}[H]
\centering
\caption[Statistical Performance Comparison of logit-based, loss-based, and ``Baseline'' Techniques Across Various Pathologies]{Statistical performance comparison between the proposed techniques logit-based and loss-based and the ``baseline'' technique across various pathologies. The upper table displays the findings of the logit-based technique, while the lower table displays the findings of the loss-based technique. The reported metrics for each pathology are the Kappa statistic, p-value, t-statistic, statistical power, Cohen's d, and Bayes Factor (BF10). A kappa value of 1 indicates perfect agreement between techniques, whereas a larger Bayes factor indicates greater support for the logit-based or loss-based technique over the baseline.}\label{tab:taxonomy.table.3.metrics}
\begin{tabular}{clrrrrrr}
    &
    & \cellcolor{table_title}{kappa}
    & \cellcolor{table_title}{p\_value}
    & \cellcolor{table_title}{t\_stat}
    & \cellcolor{table_title}{power}
    & \cellcolor{table_title}{cohen-d}
    & \cellcolor{table_title}{BF10} \\
    & Atelectasis    & 0.495 & 0 & 20.2 & 1    & 0.346 & 10   + \\
    & Consolidation  & 0.508 & 0 & 8.8  & 1    & 0.150 & 10   + \\
    & Infiltration   & 0.620 & 0 & 11.1 & 1    & 0.190 & 10   + \\
    & Edema          & 0.614 & 0 & 15.3 & 1    & 0.263 & 10   + \\
    & Pneumonia      & 0.573 & 0 & 8.2  & 1    & 0.140 & 10   + \\
    & Cardiomegaly   & 0.615 & 0 & 18.1 & 1    & 0.310 & 10   + \\
    & Lung Lesion    & 0.580 & 0 & 9.9  & 1    & 0.169 & 10   + \\
    & Lung Opacity   & 1     & 1 & 0    & 0.05 & 0     & 0.019  \\
    \multirow{-10}{*}{\begin{tabular}[c]{@{}c@{}}\\ L\\  \\ O\\ \\ G\\ \\ I\\ \\ T\end{tabular}}
    & Enlarged Cardiomediastinum & 1 & 1 & 0 & 0.05 & 0 & 0.019 \\
    \multicolumn{8}{l}{{}} \\
    &
    & \cellcolor{table_title}{kappa}
    & \cellcolor{table_title}{p\_value}
    & \cellcolor{table_title}{t\_stat}
    & \cellcolor{table_title}{power}
    & \cellcolor{table_title}{cohen-d}
    & \cellcolor{table_title}{BF10} \\
    & Atelectasis    & 0.222 & 0     & 29.3 & 1    & 0.502 & 10   + \\
    & Consolidation  & 0.310 & 0     & 23.1 & 1    & 0.396 & 10   + \\
    & Infiltration   & 0.836 & 0.053 & 1.9  & 0.49 & 0.033 & 0.125  \\
    & Edema          & 0.343 & 0     & 29.9 & 1    & 0.512 & 10   + \\
    & Pneumonia      & 0.394 & 0.207 & 1.3  & 0.24 & 0.022 & 0.043  \\
    & Cardiomegaly   & 0.501 & 0     & 21.6 & 1    & 0.370 & 10   + \\
    & Lung Lesion    & 0.059 & 0     & 31.3 & 1    & 0.537 & 10   + \\
    & Lung Opacity   & 1     & 1     & 0    & 0.05 & 0     & 0.019  \\
    \multirow{-10}{*}{\begin{tabular}[c]{@{}c@{}}\\ L\\ \\ O\\ \\ S\\ \\ S\end{tabular}}
    & Enlarged Cardiomediastinum & 1 & 1 & 0 & 0.05 & 0 & 0.019
\end{tabular}%
\end{table}

\begin{figure}[H]
    \centering
    \includegraphics[width=\textwidth]{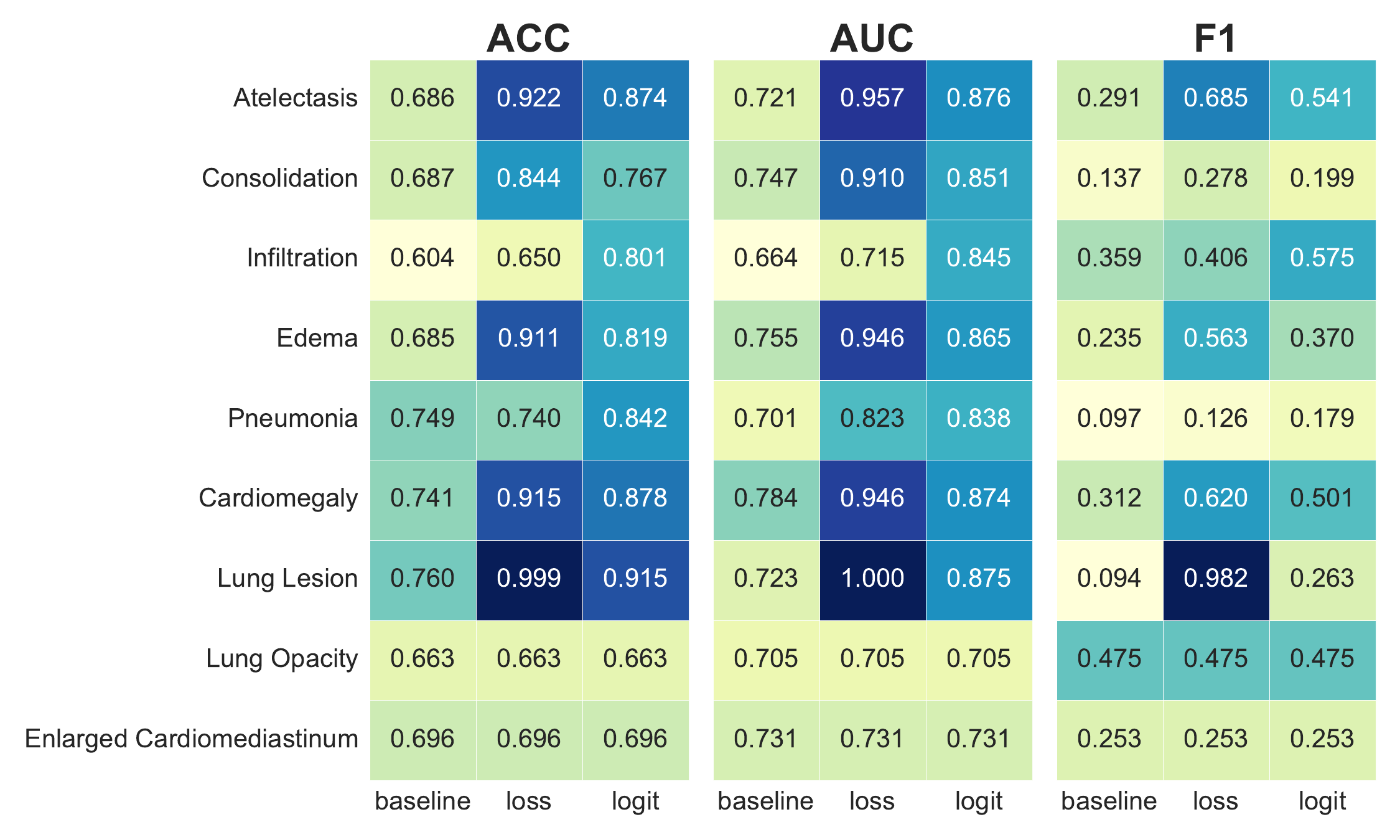}
    \caption[Heatmap Visualization of Model Performance Metrics (ACC, AUC, F1) for Different Techniques across Pathologies]{Heatmap visualization of model performance metrics across all three datasets. The subplots from left to right correspond to the Accuracy (ACC), Area Under the ROC Curve (AUC), and F1 Score for the baseline, loss-based, and logit-based techniques respectively. The pathologies are shared on the y-axis. Darker colors signify higher values, indicating better model performance. Each cell represents the value of the corresponding metric for the given technique on a specific pathology}\label{fig:taxonomy.fig.2.metrics}
\end{figure}

\begin{figure}[H]
    \centering
    \includegraphics[width=\textwidth]{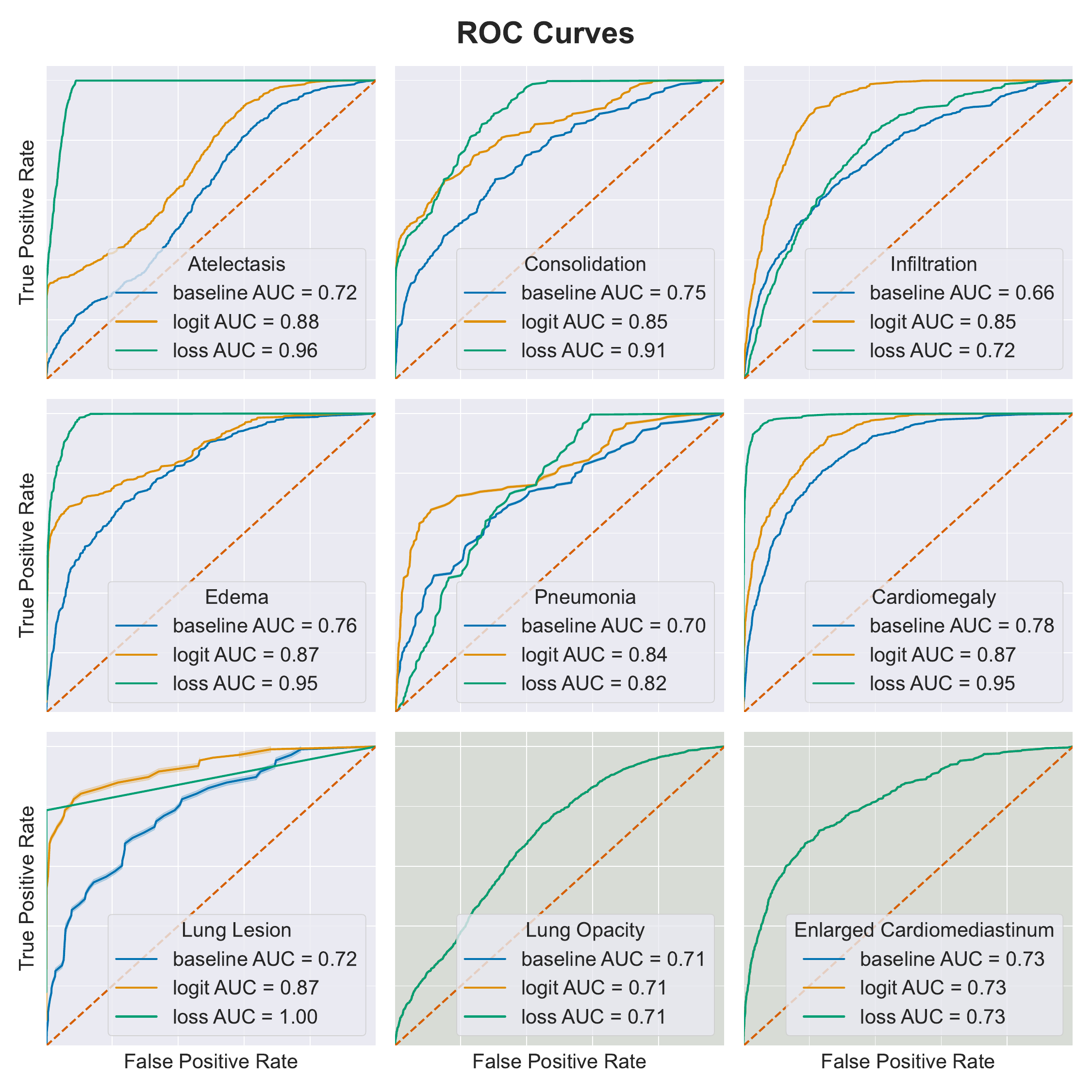}
    \caption[Comparative Analysis of ROC Curves for Nine Thoracic Pathologies: logit-based, loss-based, and ``baseline'' Techniques]{Comparative analysis of the ROC curves for nine thoracic pathologies using the logit-based and loss-based techniques as well as the baseline. The subplots highlighted with a darker background, represent parent class diseases.}\label{fig:taxonomy.fig.3.roc_curve_all_datasets}
\end{figure}

The comparative analysis presented in Figure~\ref{fig:taxonomy.fig.2.metrics} examines the performance of the proposed loss-based and logit-based methods in comparison to the ``baseline'' method across three important metrics: Accuracy (ACC), Area Under the Receiver Operating Characteristic Curve (AUC), and F1 score for different pathologies.

The loss-based and logit-based methods exhibit a distinct advantage over the ``baseline'' method in terms of accuracy. In the case of Atelectasis, the loss-based method demonstrates a notably higher accuracy of 0.922 compared to the ``baseline'' method's accuracy of 0.686. Additionally, the logit-based method achieves an accuracy of 0.874. As predicted, there is no noticeable disparity in accuracy between the methods for the parent classes, Lung Opacity and Enlarged Cardiomediastinum, as indicated by scores of 0.663 and 0.696, respectively.

The AUC, a performance measure that takes into account both sensitivity and specificity, provides further evidence of the superior performance of the loss-based and logit-based techniques. In the case of Cardiomegaly, the area under the curve (AUC) demonstrates improvements of 21\% and 11\% when employing the loss and logit techniques, respectively. The AUC values for the parent classes, Lung Opacity and Enlarged Cardiomediastinum, are consistent across all three methods.

The F1 score, which is calculated as the harmonic means of precision and recall, serves to emphasize the improved performance of our proposed methods. Significantly, in the case of Lung Lesion, the F1 score exhibits a notable increase from 0.094 in the ``baseline'' approach to 0.982 in the loss-based approach, and 0.263 in the logit-based approach.

The obtained results provides further support for our previous findings, which indicate that the utilization of the logit-based and loss-based methods leads to substantial improvements in performance compared to the ``baseline'' method across most child classes. In all measured aspects and scenarios, the loss-based method exhibits slightly superior performance compared to the logit-based method.

Table~\ref{tab:taxonomy.table.3.metrics} provides a comparative analysis of the performance of our proposed logit-based and loss-based techniques with the ``baseline'' method, using various statistical metrics. The logit-based technique, as indicated in the upper table, suggest a significant performance enhancement compared to the ``baseline'' across all evaluation tests, with kappa values ranging between 0.495 and 1. The kappa statistic is used to measure the level of agreement between two techniques, where a value of 1 signifies perfect alignment. The p-value for all child classes is below 0.05, ranging from 2.1E-89 to 2.9E-16, thereby implying a statistically significant improvement of the logit-based method over the ``baseline''. High t-statistics and power values of 1 further underscore the robustness of our technique. The Bayes factor results for the logit-based technique are exceptionally strong across all classes, suggesting substantial evidence favoring the logit-based method for these scenarios.

The proposed loss-based technique demonstrates encouraging results when benchmarked against the ``baseline'', albeit with more variability. Kappa values spanned from a minimum of 0.059 for Lung Lesion to a maximum of 0.836 for Infiltration. While the p-values indicate statistically significant improvement for most conditions, Infiltration and Pneumonia had p-values exceeding 0.05 (0.053 and 0.207, respectively), hinting that the performance improvement over the ``baseline'' for these conditions may not be statistically significant. High t-statistics and power values of 1 were observed for all conditions except Infiltration and Pneumonia. The cohen-d values for the loss-based technique were generally larger than those for the logit-based technique, signifying a larger effect size. The Bayes factor results for the loss-based technique were exceedingly strong for conditions such as Atelectasis and Edema, but considerably lower for conditions like Infiltration and Pneumonia, indicating less evidence supporting the loss-based technique for these conditions.

Both the logit-based and loss-based techniques shows considerable improvements over the ``baseline'' technique, though the degree of improvement varied. The logit-based technique exhibited a more consistent level of improvement across all conditions, whereas the loss-based technique showed potential for even larger improvements in certain conditions, albeit with less consistency across the conditions studied.
\section{Discussion and Conclusion}\label{sec:taxonomy.discussion}
In this study, we propose two novel hierarchical multi-label classification techniques, namely the loss-based and logit-based methods, to improve the accuracy and interpretability of results in applications with hierarchical class structures. The loss-based approach introduces a regularization term in the loss function, enabling a fine-grained adjustment of the hierarchical influence during model optimization. The logit-based method offers a straightforward and computationally efficient way to integrate label hierarchy by adjusting the logit outputs without extensive modifications to the existing model architecture.

The experimental results demonstrate the effectiveness of the proposed techniques in enhancing the classification accuracy of thoracic diseases across three widely-used public chest X-ray datasets: CheXpert, PADCHEST, and NIH. The substantial improvements in various performance metrics, including accuracy, AUC, F1 scores, Cohen's d, Cohen's kappa, t-statistics, p-value, and Bayes factor, compared to the baseline model, highlight the robustness and reliability of our methods. These findings suggest that the loss-based and logit-based techniques can serve as valuable tools for improving multi-label classification performance while providing a higher level of interpretability by leveraging the hierarchical relationships among classes.

The proposed techniques harness the disease taxonomy to enhance classification performance, emphasizing the importance of incorporating label relationships in classification tasks. By providing predictions at varying levels of granularity based on the taxonomy, these hierarchical techniques could assist healthcare professionals in making more accurate and personalized diagnoses. Moreover, the integration of these methods into computer-aided diagnosis systems has the potential to streamline the diagnostic process, reduce the workload of clinicians, and ultimately improve patient outcomes. The taxonomy structure presented in Figure~\ref{fig:taxonomy.fig.1.taxonomy_structure} plays a crucial role in the effectiveness of our proposed hierarchical multi-label classification techniques. By organizing lung pathologies in a hierarchical manner, the taxonomy enables our models to capture the inherent relationships between different abnormalities. This structured representation of knowledge allows the models to make more informed predictions, as they can leverage the contextual information provided by the parent-child relationships. For example, when classifying a specific condition such as consolidation, the model can take into account the presence or absence of its parent categories, such as lung opacity and atelectasis, to make a more accurate prediction. This hierarchical approach not only improves the overall classification performance but also enhances the interpretability of the results, as the predictions are made within the context of the disease hierarchy. The integration of domain knowledge through the taxonomy structure is a key strength of our proposed techniques, setting them apart from traditional flat classification approaches that treat each pathology independently.

While existing methods mentioned in the related work section have made notable contributions to hierarchical classification in various domains, our research specifically focuses on addressing the challenges of multi-class problems in medical imaging, an area that has received less attention. The loss-based and logit-based techniques introduce novel approaches to incorporate label hierarchy directly into the model's optimization process or output adjustment, offering methodological differences that make a direct comparison with existing methods challenging and beyond the scope of this paper. Furthermore, the logit-based approach's computational efficiency makes it more adaptable to real-world scenarios with limited resources, a key advantage over many existing methods that rely on complex architectures or training schemes.

The generalizability and effectiveness of the proposed techniques are demonstrated through extensive evaluations on three large-scale, diverse chest X-ray datasets, highlighting their potential for real-world application. Although a direct comparison with existing methods could yield valuable insights, the primary focus of this paper is to introduce and validate the proposed techniques in the specific context of chest radiography. A comparative analysis with existing methods could be an interesting direction for future research, but the current study aims to establish the effectiveness and novelty of our proposed techniques in the domain of multi-label classification for medical imaging.

However, there are some limitations to these methods that should be acknowledged. Applying these techniques to other applications would require the development of a taxonomical structure for the dataset labels, which can be challenging for complex applications and may require consensus among domain experts. Additionally, the effectiveness of the proposed techniques could be influenced by the quality and consistency of dataset labeling, which may vary across different sources. Future research should focus on evaluating these techniques across a broader range of datasets and investigating the impact of labeling quality on performance.

In conclusion, the loss-based and logit-based hierarchical multi-label classification techniques introduced in this study demonstrate significant improvements in the accuracy and interpretability of thoracic disease classification from chest X-ray images. By leveraging the hierarchical relationships among classes, these methods offer a promising approach to enhance the performance of multi-label classification models in medical imaging applications. Further research is necessary to explore their potential benefits in clinical settings and to address the limitations associated with taxonomy development and labeling quality. Nonetheless, the proposed techniques represent a valuable contribution to the field of multi-label classification in medical imaging and have the potential to improve diagnostic accuracy and patient care.

\bibliographystyle{bst/elsarticle-num}
\bibliography{Better_BibTeX_Zotero}

\end{document}